\documentclass{sage1}

\usepackage{url}
\usepackage{IEEEtrantools}

\usepackage[numbers,sort&compress]{natbib}
\bibpunct[, ]{(}{)}{,}{a}{}{,}
\RequirePackage{amsmath}
\RequirePackage{amsthm}
\RequirePackage{xspace}
\usepackage{graphicx}
\usepackage{color}
\usepackage{amsmath,amssymb,amsthm}
\usepackage{algorithm}
\usepackage{algpseudocode}
\usepackage{subfig}
\usepackage{listings}
\usepackage{stmaryrd} % for \llbracket, \rrbracket
\usepackage{calc}

\usepackage{soul}

\newcommand{\intco}[1]{\ensuremath{{\left[#1\right)}}}

\DeclareMathOperator{\FF}{\rotatebox[origin=c]{45}{$\Box$}}
\DeclareMathOperator{\XX}{\bigcirc}
\DeclareMathOperator{\GG}{\Box}

\newcommand{\num}[1]{\relax\ifmmode \mathbb #1\else $\mathbb #1$\fi}
\newcommand{\nnnum}[1]{\relax\ifmmode 
  {\mathbb #1}_{\geq 0} \else ${\mathbb #1}_{\geq 0}$
  \fi}
\newcommand{\npnum}[1]{\relax\ifmmode 
  {\mathbb #1}_{\leq 0} \else ${\mathbb #1}_{\leq 0}$
  \fi}
\newcommand{\pnum}[1]{\relax\ifmmode 
  {\mathbb #1}_{> 0} \else ${\mathbb #1}_{> 0}$
  \fi}
\newcommand{\nnum}[1]{\relax\ifmmode 
  {\mathbb #1}_{< 0} \else ${\mathbb #1}_{< 0}$
  \fi}
\newcommand{\plnum}[1]{\relax\ifmmode 
  {\mathbb #1}_{+} \else ${\mathbb #1}_{+}$
  \fi}
\newcommand{\nenum}[1]{\relax\ifmmode 
  {\mathbb #1}_{-} \else ${\mathbb #1}_{-}$
  \fi}

\newcommand{\reals}{{\num R}}                    %reals
\newtheoremstyle{mytheoremstyle} % name
        {\topsep}                    % Space above
        {\topsep}                    % Space below
        {\rm\fontfamily{ptm}\selectfont}                   % Body font
        {}                           % Indent amount
        {\fontfamily{ptm}\selectfont\bf}                   % Theorem head font
        {.}                          % Punctuation after theorem head
        {.5em}                       % Space after theorem head
        {}  % Theorem head spec (can be left empty, meaning ‘normal’)
\theoremstyle{mytheoremstyle}

\newtheorem{definition}{Definition}

\newcounter{rem}
\newtheorem{example}{Example}
\newtheorem{problem}{Problem}
\def\A{{\cal A}} % HA
\def\G{{\cal G}} % pieces of SHIOA
\def\R{{\cal R}} % relation
\def\U{{\cal U}} % set of trajectories
\def\V{{\cal V}} % Lyapunov function
\def\W{{\cal W}}
\def\X{{\cal X}} % Lyapunov function
\def\Y{{\cal Y}} % set of trajectories
\usepackage{xspace}

\newcommand{\true}{${\tt True}$\xspace}
\newcommand{\false}{${\tt False}$\xspace}

\journal{The International Journal of Robotics Research}
\volume{000}
\issue{00}
\copyrightline{$\copyright$ Copyright}
\firstpage{1}
\lastpage{13}
\doi{doi number}
\articletype{Article type}
\pubyear{2010}
\begin{document}

\title{Dynamics-Based Reactive Synthesis and Automated Revisions for High-Level Robot Control}
%\title{Working title: User-in-the-Loop Refinements to Reactive Specifications for Dynamical Systems}

% use a multiple column layout for up to three different affiliations
% You will get a Paper-ID when submitting a pdf file to the conference system

\author{Jonathan A. DeCastro
\thanks{This work was supported by the NSF Expeditions in Computing project ExCAPE: Expeditions in Computer Augmented Program Engineering [grant number CCF-1138996].}
\thanks{Corresponding author; e-mail: jad455@cornell.edu}
}
\address{Sibley School of Mechanical and Aerospace Engineering, Cornell University, Ithaca, NY 14853, USA}

\author{R\"{u}diger Ehlers}
\address{Department of Computer Science, University of Bremen, 28359 Bremen, Germany}

\author{Matthias Rungger}
\address{Department of Electrical Engineering and Information Technology, Technical University of Munich, 80333 Munich, Germany}

\author{Ay\c{c}a Balkan and Paulo Tabuada}
\address{Electrical Engineering Department, University of California, Los Angeles, Los Angeles, CA 90095, USA}

\author{Hadas Kress-Gazit}
\address{Sibley School of Mechanical and Aerospace Engineering, Cornell University, Ithaca, NY 14853, USA}

\maketitle

\begin{abstract}
%\boldmath
The aim of this work is to address issues where formal specifications cannot be realized on a given dynamical system subjected to a changing environment.  Such failures occur whenever the dynamics of the system restrict the robot in such a way that the environment may prevent the robot from progressing safely to its goals.  We provide a framework that automatically synthesizes revisions to such specifications that restrict the assumed behaviors of the environment and the behaviors of the system. We provide a means for explaining such modifications to the user in a concise, easy-to-understand manner.  Integral to the framework is a new algorithm for synthesizing controllers for reactive specifications that include a discrete representation of the robot's dynamics. The new approach is demonstrated with a complex task implemented using a unicycle model.
\end{abstract}

\section{Introduction} \label{sec:intro}
One of the benefits of the formal methods approach to robot mission planning is that it frees users from the burdens of programming controllers for complicated robot tasks, removing the arduous step of re-validation every time the task changes; see, e.g.~\cite{tabuada2006,KGFP_TRO09,Wongpiromsarn2010,Kloetzer2008,wolff2013}.  Application of provably-correct controllers on fielded systems requires that the correctness guarantees extend to a possibly complex set of dynamics describing the robot.  Recently, tools have been proposed for generating discrete abstractions for a wide class of nonlinear systems, e.g.~\cite{PGT08,girard2010,Rei11,ZPMT12}.  It is natural to exploit such abstractions when synthesizing controllers that are applicable to complex physical systems.  This paper serves to solidify such a synthesis approach.

When composing tasks for applications such as self-driving cars, household robots, and disaster response robots, the user must specify both the desired system behaviors and any assumptions on the sensed environment.  Often, such assumptions incorporate knowledge of the behavior of humans and other elements in the robot's physical surroundings.  If any of these assumptions conflict with the specified task in the context of a robot's dynamics, the specification is considered to be {\em unrealizable}.  That is to say, the environment is allowed to work against fulfilment of the specification by the robot assigned to the task.  As the underlying cause for unrealizability is often difficult to determine, automated frameworks for debugging specifications (\cite{KonighoferHB09,raman13}) and generating environment assumptions (\cite{fainekos11,LiDS11,AlurMT13}) have gained attention recently.

This paper seeks to address the problem of realizability of specifications for robots with dynamics by introducing a novel framework that {\em automatically} suggests additions to the specification to resolve unrealizability issues.  We do so by introducing a new synthesis approach that takes a specification agnostic to the dynamics of the robot, and synthesizes a controller (if possible) for a specific robot, given its discrete abstraction.  If it turns out that the specification is unrealizable, we automatically generate revisions to the specification that render the task realizable with respect to the robot dynamics.  These formulas effectively alter the environment assumptions and robot behaviors.  The drawback of generating revisions independently of the user is that the resulting specifications may fail to capture the original intent of the mission.  We address this by advocating for an means explaining such revisions in a manner that is simple to understand, permitting user interaction as the revisions are computed. 

When considering dynamics, unrealizability can arise as a result of unwanted effects such as overshooting into undesired regions or an inability to make progress toward goals.  Take the following scenario where rooms A and B are separated by a door:
\begin{center}
\textit{Visit rooms A and B.  Avoid the door if it is closed.  Assume that, infinitely often, the door is open.}
\end{center}
In the above scenario, the robot must be able to robustly avoid the walls of the workspace (boundaries) and appropriately react to the door as it opens and closes.  If one applies this specification to robot with inertia, the task could fail because the robot may not be able to stop by the time the door closes, violating the statement {\em ``Avoid the door if it is closed.''}  By leveraging both the system dynamics and the specification, we could recover by adding the environment assumption \textit{``The door must not close if within 1 meter of it.''}  In this paper, we explore an automated method for suggesting such revisions.

The remainder of this paper is outlined as follows.  In Section~\ref{sec:setup}, we review relevant formal definitions and notation.  We formally state the problem in Section~\ref{sec:prob}, and address related work in the context of the problem in Section~\ref{sec:related}.  In Section~\ref{sec:synth}, we describe our approach for synthesis for specifications involving discrete abstractions.  We present the approach for generating formulas and user feedback in Section~\ref{sec:envSafetyAssump}.  Next, we demonstrate our approach for a complex task carried out by an abstraction of a unicycle robot model in Section~\ref{sec:example}.  Lastly, we provide a summary in Section~\ref{sec:conclusion}.

\section{Preliminaries} \label{sec:setup}

%In this section, we introduce the system models of the robot and the environment, and present the reactive controller design procedure from linear temporal logic specifications. 

%\JDC{mention that this is close to the semantics in [Liu, Ozay], and so we do not stray much from usual semantics. Therefore, our techniques can be applied to many different semantics, such as those.}

\subsection{Linear Temporal Logic}
Linear temporal logic (LTL) formulas are defined over the set $AP$ of atomic propositions in the recursive grammar:
\[
\varphi ::= true \mid \pi \mid \varphi_1 \wedge \varphi_2 \mid \neg\varphi \mid \XX\varphi \mid \varphi_1 \mathsf{U} \varphi_2
\]
where $\pi$ is an atomic proposition in $AP$.  Respectively, $\wedge$ and $\neg$ are the Boolean operators ``conjunction'' and ``negation'', and $\XX$ and $\mathsf{U}$ are the temporal operators ``next'' and ``until''.  From these operators, the following operators are derived: ``disjunction'' $\vee$, ``implication'' $\Rightarrow$, ``equivalence'' $\Leftrightarrow$, ``always'' $\GG$, and ``eventually'' $\FF$.    

$AP$ consists of a set of {\em environment} propositions $\X$
%JDC{move to later section: $ = \{P_{e,1},\ldots,P_{e,p_e}\}$} 
describing the state of the discretized robot sensor values and a set of {\em system} propositions $\Y$ describing the current pose of the robot in the workspace.  For a set of propositions $\X$, let $\XX\X = \{\XX \pi_i\}_{\pi_i\in\X}$ be the set of propositions with the ``next'' operator.  The LTL formulas are evaluated over infinite sequences $\sigma = \sigma_0 \sigma_1 \sigma_2 \ldots$ of truth assignments to the propositions in $AP$.  $\sigma$ is said to {\em satisfy} $\XX\varphi$, $\GG\varphi$, or $\FF\varphi$ if $\varphi$ holds true in the next position in the sequence, every position, or at some future position(s), respectively.  We refer the reader to \cite{Var96} for a complete definition of the syntax and semantics of LTL formulas.

%We say that \emph{the robot $S_r(\tau)$ satisfies $\varphi$ with respect to the environment $S_e$} if any behavior of $\rho_r\in B(S_r(\tau))$ and \mbox{$\rho_e\in B(S_e)$} satisfy $\varphi$. 

\begin{definition}[Robot Mission Specification] 
The specifications we consider are LTL formulas of the form:
\begin{IEEEeqnarray*}{rCl}
\varphi &:=& \underbrace{(\varphi_i^e \wedge \varphi_t^e \wedge \varphi_g^e)}_{\varphi^e} \implies \underbrace{(\varphi_i^s \wedge \varphi_t^s \wedge \varphi_g^s)}_{\varphi^s}
\end{IEEEeqnarray*}
The formulas $\varphi_i^\alpha$, $\varphi_t^\alpha$, and $\varphi_g^\alpha$ are defined over $AP$, where $\varphi_i^\alpha$ are formulas for the initial conditions, $\varphi_t^\alpha$ the allowed transitions (safety conditions) to be satisfied always, $\varphi_g^\alpha$ the goals (liveness conditions) to be satisfied infinitely often, and $\alpha=\{e,s\}$ (with $e$ for `environment' and $s$ for `system').  The liveness guarantees take the form $\bigwedge_{i \in I_\alpha}\GG\FF(B_{i}^{\alpha})$, where $I_\alpha$ is the index set of environment goals ($B_{i}^e$) or system goals ($B_{i}^e$).
\end{definition}
%\JDC{make note of the $\rho$s and their arguments here if getting into those details later}
%\JDC{Note here that the environment transitions in $S_e$ manifest as conditions applied to $\varphi_t^e$.}

\begin{definition}[Controller Strategy and Execution] 
\label{d:strategy}
Define a {\em controller} as a finite-state machine $\A = (S,S_0,\X,\Y,\delta,\gamma_\X,\gamma_\Y)$, where $S$ is the set of controller states, $S_0\subseteq S$ is the set of initial controller states, $\X$ and $\Y$ are sets of propositions described above, $\delta : S \times 2^{\X} \to S$ is a state transition relation providing the next state $s'\in S$ given the next state $s\in S$ and the current value of the environment input $z \in 2^{\X}$ , i.e. $s' = \delta(s,z)$, $\gamma_\X : S \to 2^\X$ is a labelling function mapping controller states to the set of environment propositions evaluating to \true for all transitions into that state, and $\gamma_\Y : S \to 2^{\Y}$ is a labelling function mapping controller states to the set of robot configuration propositions evaluating to \true in that state.

Consider an infinite {\em execution} $\sigma$ of $\A$, where $\sigma = (\gamma_\X(s_0),\gamma_\Y(s_0))(\gamma_\X(s_1),\gamma_\Y(s_1)),\ldots$ for $s_0 \in S_0$ and $s_i\in S$, $i > 0$.  $\varphi$ is deemed {\em realizable} if there exists a finite-state machine $\A$ such that every execution produced by $\A$ satisfies $\varphi$.  That is, at every $i \geq 0$, there exists an assignment of system variables $\Y$ for all possible assignments of the environment variables $\X$ such that $\sigma$ satisfies $\varphi$.  If there exist some environment behaviors on $\X$ for which no such $\A$ can be found, then $\varphi$ is {\em unrealizable}.  
\end{definition}
%We declare a specification to be {\em trivial} if there exists a finite-state machine $\A$ that 
A discrete controller can be synthesized from a specification follows by solving a two-player game played between the environment and the system, as described in~\cite{BJPPS12}.  

%The types of specifications we consider are {\em reactive}; that is, the actions of the robot depend on the sensed environment inputs.  

%\JDC{We declare a specification to be {\em trivially realizable} if it...}
%\JDC{A {\em counterexample} is defined to be a set of atomic proposition truth-values for which $\varphi$ is rendered unrealizable.}

\subsection{Synthesis over a Connectivity Graph}
A connectivity graph is an undirected graph describing those workspace regions that are accessible and adjacent to one another.
%The controller that is generated may be applied to the robot assuming that a separate motion planning scheme can be invoked to satisfy the connectivity constraints.
%For a two-dimensional workspace into a uniform grid, the topological connectivity is advantageous for composing specifications, as the actions are simple to interpret.  For example, actions such as ``stay in the current room'' or ``move North'' can be expressed by repeated application of a single, simple control actions.  

Given a bounded configuration space $W\subset\num{R}^n$, let $\R = \{R_{1}\ldots R_{p}\}$ represent a set of disjoint regions whose closure covers $W$, where the open sets $R_i\subseteq W$.  Let $\Y$ be the set of propositions representing the workspace regions over which the topology model will be specified and $\pi_{i}\in\Y$ denote a region proposition that evaluates to \true when the robot is in $R_i\in\R$.  
\begin{definition}[Topology Model] 
We define the {\em topology model} as a formula $\varphi_t^{top}$ over $\Y$ that encodes the allowed next regions given the current region.  This formula is defined as follows:
\[
\varphi_t^{top} =  \bigwedge_{\pi_{i}\in\Y} \GG \left( \pi_{i}\implies \bigvee_{\substack{\pi_{j}\in\Y : \\ cl(R_i) \cap cl(R_j) \neq \varnothing}} \XX \pi_{j} \right),
\]
where $cl(\cdot)$ denotes the closure operation on a set.  Note that we also enforce mutual exclusion of regions; that is, the robot is only allowed to occupy one region at a time.
\end{definition}
%\JDC{todo: adjacent corners are allowed in this definition}

% and $\Y=\{P_{1}\ldots P_{p}\}$ represent the set of propositions, for which some $P_i$ holds true when the robot is physically inside the region labeled $R_i$.  
%Under the region propositions we define a {\em topological connectivity model} $S_{top}$ as the tuple $(Q_{top},V_{top},\delta_{top},\gamma_{top})$, where:
%\begin{itemize}
%  \item $Q_{top} = \{1\ldots p\}$ is the set of states with each $q_i\in Q_{top}$ representing the region $R_i\in\R$;
%  \item $V_{top} = \{1\ldots p\}$ is the set of robot locomotion commands representing the desired post state;
%  \item $\delta_{top} : Q_{top} \to 2^{V_{top}}$ are the set of allowed transitions defined such that, for some $q_i\in Q_{top}$, $q_j\in V_{top}$, $q_j \in \delta_{top}(q_i)$ iff $cl(R_i) \cap cl(R_j) \neq \varnothing$, where $cl(\cdot)$ denotes the closure operation on a set;
%  \item $\gamma_{top} : Q_{top} \to \Y_{top}$ labels each discretized configuration with the associated proposition in $\Y_{top}$ that evaluates to \true.
%\end{itemize}

When combining the user specifications $\varphi^e$ and $\varphi^s$ with the topological model, we obtain the formula $\varphi^e \implies (\varphi^s \wedge \varphi_t^{top})$ written over $AP^{top} = \X \cup \Y$.  A controller $\A$ is synthesized if $\varphi^e \implies (\varphi^s \wedge \varphi_t^{top})$ is realizable.

\subsection{Synthesis over Robot Abstractions}
\label{s:synthAbstr}
Our model of the behavior of the robot is a nonlinear differential equation
\begin{IEEEeqnarray}{c'c}\label{e:sys}
  \dot\xi(t) = f(\xi(t),\nu(t))
\end{IEEEeqnarray}
given by the function $f:\num{R}^n\times\num{R}^m \to \num{R}^n$, where $\xi(t)$ is the continuous state of the robot and $\nu(t)$ the command input of the robot at time $t\in\num{R}_{\ge0}$.  We use $\xi_{x,\nu}$ to denote a \emph{trajectory} of~\eqref{e:sys} with initial state $x$ and input signal $\nu$.  We impose the usual regularity assumptions on $f$ that imply the existence and uniqueness of solutions of~\eqref{e:sys}.  Moreover, we assume that~\eqref{e:sys} is forward-complete (its solution is defined for all $t\geq 0$), see e.g.~\cite{AS99}.

%A \emph{behavior $\rho_r$ of $S_r(\tau)$} is defined as usual as an infinite sequence of states, for which there exists an infinite sequence $\nu_r$ of inputs, such that the two sequences satisfy the transition relation $(\rho_{r,i},\nu_{r,i},\rho_{r,i+1})\in \underset{r}{\lto}$for all times $i\in\num{N}$. We use $B(S_r(\tau))$ to denote the set of all behaviors of $S_r(\tau)$. \JC{behaviors: keeping?}

We produce an estimate of the divergence of neighboring trajectories to obtain a conservative over-approximation of the continuous dynamics operating under a sample-and-hold controller framework.  Let $\beta:\num{R}_{\ge0}\times\num{R}_{\ge0}\to\num{R}_{\ge0}$ be a continuous function, that satisfies $\beta(0,0)=0$ and $\beta(\cdot,t)$ is strictly increasing for every $t\in\num{R}_{\ge0}$.  Let us assume that any pair of trajectories produced by the robot dynamics~\eqref{e:sys} satisfy 
\begin{IEEEeqnarray}{c}\label{e:fc}
  |\xi_{x,\nu}(t) - \xi_{x',\nu}(t) | \le \beta(|x - x'|,t),
\end{IEEEeqnarray}
with $x,x'\in \num{R}^n$, $t\in\num{R}_{\ge0}$ and constant input function $\nu\in U$.  Given that~\eqref{e:sys} is Lipschitz continuous with constant $L$, the function $\beta(r,t) = re^{Lt}$ satisfies~\eqref{e:fc}.

In constructing a discrete abstraction, we follow the approach in~\cite{Rei11,PGT08,ZPMT12} for discretizing the bounded configuration space $W\subset\reals^n$ and the bounded space of command inputs $U\subset\reals^m$.  We denote $[W]_\eta$ to be the uniform grid on $W$ discretized with resolution $\eta$.  This grid is defined as follows:  
\begin{IEEEeqnarray*}{c}
  [W]_\eta := \{x\in W\mid \exists k\in\num{Z}^n: x = k\eta\}.
\end{IEEEeqnarray*}
We introduce the discretization parameters $\tau,\eta,\mu\in\reals_{>0}$, where $\eta$ and $\mu$ define, respectively, the discretization of the robot configuration and command input spaces, and $\tau$ represents the sampling time.  

We next denote a set of {\em configuration} propositions $\Y_a$ describing the robot's pose, and a set of {\em locomotion command} propositions $\U_a$
%\JDC{$ = \{P_{\nu,1},\ldots,P_{\nu,p_\nu}\}$} 
describing the controlled actions taken by the robot.  We assume throughout that $\U_a$ consists of purely robot locomotion commands; the set can be generalized to capture other robot actions (e.g. wave hand, activate alarm), but this extension is omitted in this paper.  
\begin{definition}[Discrete Abstraction] 
\label{d:abstraction}
The {\em discrete abstraction} $S_a(\tau,\eta,\mu)$ is defined by the tuple $(Q_a,V_a,\delta_a,\gamma^\Y_a,\gamma^\U_a)$, where:
\begin{itemize}
  \item $Q_a = [X]_\eta$ is the set of discretized robot configurations;
  \item $V_a = [U]_\mu$ is the set of discretized robot locomotion commands;
  \item $\delta_a : Q_a \times V_a \to 2^{Q_a}$ is a transition relation defined such that, there exists $q_a,q_a'\in Q_a$ and $v_a\in V_a$, $q_a'\in\delta_a(q_a,v_a)$ iff there exists $\xi_{x,\nu}$ with $\nu(t) = v_a$, for all $x \in \{x_0 \mid |x_0 - q_a| \le \eta/2\}$ and $t\in\intco{0,\tau}$, such that 
  \begin{IEEEeqnarray}{c}\label{e:abstransFC}
    |q_a' - \xi_{x,\nu}(\tau)| \le \beta(\eta/2,\tau) + \eta/2;
  \end{IEEEeqnarray}
  \item $\gamma^\Y_a : Q_a \to \Y_a$ labels each discretized configuration with the associated proposition in $\Y_a$ that evaluates to \true when the robot is in the given configuration;
  \item $\gamma^\U_a : V_a \to \U_a$ labels each discretized command with the associated proposition in $\U_a$ that evaluates to \true when the given command is active.
\end{itemize}
Additionally, we define $\gamma_a : \R \to 2^{\Y_a}$, a labelling function associating each region to a set of configuration propositions, as $\gamma_a(R_j) = \{\gamma^\Y_a(q_a) \mid q_a\in [R_j]_\eta\}$.  Finally, we encode the transitions $\delta_a$ as a safety formula $\varphi_t^a$ defined over $\Y_a\cup\XX\U_a\cup\XX\Y_a$ of the form
\begin{IEEEeqnarray*}{rCl}
\varphi_t^a = \bigwedge_{\substack{\gamma^\Y_a(q_a)\in\Y_a \\ \gamma^\U_a(v_a)\in\U_a}} \GG  \left( 
\vphantom{\bigvee_{\substack{\gamma^\Y_a(q_a')\in\Y_a : \\ q_a'\in\delta_a(q_a,v_a)}}} 
\left( \vphantom{\gamma^\Y_a(q_a)} \right. \right. &\gamma^\Y_a(q_a)& \left. \left. \wedge \XX\gamma^\U_a(v_a) \right) \implies \bigvee_{\substack{\gamma^\Y_a(q_a')\in\Y_a : \\ q_a'\in\delta_a(q_a,v_a)}} \XX \gamma^\Y_a(q_a') \right) .
\end{IEEEeqnarray*}
\end{definition}
Suppose $\Y_a = \{\pi_0,\pi_1,\pi_2\}$, $\U_a = \{\pi_{fwd}\}$ and we start at grid cell $\pi_0$ at $q_0$.  Assume that the set of possible successor configurations under this command consists of cells 1 and 2; i.e. $\delta_a(q_0,v_0) = \{q_1,q_2\}$, where $\gamma^\Y_a(q_1) = \pi_1$, $\gamma^\Y_a(q_2) = \pi_2$ and $\gamma^\U_a(v_0) = \pi_{fwd}$.  In this example, we write $\GG\left((\pi_0 \wedge \XX\pi_{fwd}) \implies \XX(\pi_1 \vee \pi_2)\right)$ as the corresponding safety formula.
%\JDC{also note that since $\tau$ is the same for $S_d$ and $S_a$, we preserve the next operator and the semantics of the environment (it can only change once every $\tau$, though this isn't really mentioned in the LTL section)}

\subsection{Ensuring Correctness of the Continuous Behaviors}
\label{s:synthCorrectContinuous}
In the following, we briefly review the rules derived in~\cite{FGKGP09,FP09,liu14,liuTOM12} for modifying LTL formulas $\varphi^e$, $\varphi^s$ on $AP^{top}$ to account for the behaviors of the continuous system in between sampling instants.
%The types of region inflation rules we adopt are due to [Fainekos, Liu], while the robustness computations are based on the work of [Pola?].  
Given the system \eqref{e:sys}, there exists a constant $b\in\num{R}_{\ge0}$ such that for all $\xi(t)\in W$, $u\in U$ and $t\in\intco{0,\tau}$, we have
\begin{IEEEeqnarray}{c't'c}\label{e:bound}
|f(\xi(t),u)|\le b.
\end{IEEEeqnarray}
%Here we use $+$ to denote the Minkowski set addition and $\num{B}$ to denote the closed ball in $\num{R}^n$ with respect to the infinity
%norm $|\cdot|$ with radius $\tau b$. 
%In case that~\eqref{e:sys} is bounded, we can simply take $W'=\num{R}^n$.

Through the process of inflation and deflation of physical workspace regions $\R$, we obtain the {\em $S_a$-strengthened} formulas $\hat\varphi^e$ and $\hat\varphi^s$.  Conservative under-approximations $\check R^\varepsilon_{i}$ (where regions are deflated or shrunk) and over-approximations $\hat R_{i}^\varepsilon$ (where regions are inflated) may be constructed as follows:
%\mbox{$\Y_r^\varepsilon=
%\{\check P_{r,i}^\varepsilon\}_{i\in\intcc{1;p_r}}\cup\{\hat
%P_{r,i}^\varepsilon\}_{i\in\intcc{1;p_r}}$} of modified atomic propositions
%by
\begin{IEEEeqnarray*}{rCl}
\check R^\varepsilon_{i} &=& \{x\in \reals^n\mid \{x\} + \varepsilon\num{B}\subseteq R_{i}\}\\
\hat R^\varepsilon_{i} &=& \{x\in \reals^n\mid x\in R_{i} + \varepsilon\num{B}\}.
\end{IEEEeqnarray*}
%We now introduce the new region definitions $\R^\varepsilon = \{\check R_{i}^\varepsilon\}_{i\in\intcc{1;p_r}} \cup \{\hat R_{i}^\varepsilon\}_{i\in\intcc{1;p_r}}$. 
Here we use $+$ to denote the Minkowski set addition and $\num{B}$ to denote the closed ball in $\num{R}^n$ with respect to the infinity norm $|\cdot|$ with radius $\tau b$.  Supposing that the formulas $\varphi^e$, $\varphi^s$ are in \emph{negated normal form}, i.e., the negation only appears in front of atomic propositions, we define the $S_a$-strengthened formulas $\hat\varphi^e$, $\hat\varphi^s$ by
\begin{IEEEeqnarray}{c}\label{e:epsformula}
  \hat\varphi^\alpha
  =
  \varphi^\alpha[\neg\pi_{top,i} / \neg\gamma_a(\hat R_{i}^\varepsilon),
  \pi_{top,i} / \gamma_a(\check R_{i}^\varepsilon)],
\end{IEEEeqnarray}
in which we are replacing the region propositions $\pi_{top,i} \in \Y$ with $\gamma_a(\check R^\varepsilon_{i})$ and $\neg\pi_{top,i}$ with $\neg\gamma_a(\hat R^\varepsilon_{i})$. 

By application of Theorem 1 of~\cite{liu14} to the semantics in~\cite{FGKGP09}, it is shown that, by choosing $\varepsilon \ge b\tau$, we are assured that the continuous trajectory of \eqref{e:sys} satisfies $(\varphi^e \wedge \varphi_t^a) \implies \varphi^s$ if the sampled-time execution satisfies $(\hat\varphi^e \wedge \varphi_t^a) \implies \hat\varphi^s$.  

Similar to~\cite{liu2013}, we treat the possible non-determinism in the abstraction as adversarial: at each time-step, the environment assigns values to the environment $\X$.  Then, the locomotion command is updated based on this input, and the abstraction chooses potential successor states to transition into given the current locomotion command.  We therefore treat $\varphi_t^a$ as a statement on the environment and adopt the formula $(\hat\varphi^e \wedge \varphi_t^a) \implies \hat\varphi^s$ over $AP = \X \cup \Y_a \cup \U_a$.  We describe in Section~\ref{sec:synth} our procedure for checking realizability for this formula. 

The synthesis of a discrete controller $\A_{S_a}$ given the formula $\varphi^e \implies (\varphi^s \wedge \varphi_t^{top})$ and an abstraction $S_a$ involves the following steps: (1) computation of the formula $\varphi^a_t$ from $S_a$ (Section~\ref{s:synthAbstr}); (2) computation of the $S_a$-strengthened formulas $\hat\varphi^e$, $\hat\varphi^s$ from the original formulas $\varphi^e$, $\varphi^s$ (Section~\ref{s:synthCorrectContinuous}); and (3) extracting a strategy as is discussed in Section~\ref{sec:synth}.

\section{Problem Formulation} \label{sec:prob}
Let $\varphi$ be a {\em realizable} formula, defined as
\[
\varphi := \varphi^e \implies (\varphi^s \wedge \varphi_t^{top}),
\]
in the set of atomic propositions $AP^{top}$ under the assumption of a topological model $\varphi_t^{top}$, where the formulas $\varphi^e$ and $\varphi^s$ are, respectively, the user-defined environment assumptions and system guarantees.  Additionally, we require that $\varphi^e$ is not falsified by robot behaviors satisfying $\varphi^s$ (i.e. no trivial behaviors).  We refer to $\varphi$ as a {\em general} formula.  The goal is to synthesize controllers for such specifications.

\begin{problem}[Synthesis under non-deterministic abstractions]
\label{pr:probSynth}
Given the subformula $\varphi_t^a$ for the robot abstraction and the $S_a$-strengthened formulas $\hat\varphi^e$ and $\hat\varphi^s$, synthesize a controller for the robot-specific specification
\[
\varphi^{abs} := (\hat\varphi^e \wedge \varphi_t^{a}) \implies \hat\varphi^s .
\]
in the set of propositions $AP$.
\end{problem}

If $\varphi^{abs}$ is {\em unrealizable}, determine a set of revisions to the specification that render the new specification realizable.  A revision is a temporal logic formula that may be conjuncted with the original formula.  Recall $\varphi^{abs}$ is unrealizable if there exists some environment behavior admissible by the formula $\hat\varphi^e \wedge \varphi_t^{a}$ such that no system behaviors satisfy $\hat\varphi^s$.

\begin{problem}[Generating revisions]
\label{pr:prob1}
Given a realizable $\varphi$, an unrealizable $\varphi^{abs}$, an abstraction $S_a$, and the $S_a$-strengthened formulas $\hat\varphi^e$ and $\hat\varphi^s$, automatically derive a set of formulas for the environment and the system such that
\begin{IEEEeqnarray}{lr}
\label{e:varphimod}
\varphi^{mod} := (\hat\varphi^e \wedge \varphi_t^a \wedge \psi_g^e \wedge \psi_t^e) \implies (\hat\varphi^s \wedge \psi_t^s)
\end{IEEEeqnarray}
is realizable and $\hat\varphi^e \wedge \varphi_t^a \wedge \psi_g^e \wedge \psi_t^e$ and $\hat\varphi^s \wedge \psi_t^s$ are satisfiable.
\end{problem}
If such formulas exist, the user is provided with a suggested liveness assumption $\psi_g^e$ and suggested set of safety assumptions and guarantees, $\psi_t^e$ and $\psi_t^s$, respectively.  For each transition formula, a simple-to-interpret statement summarizing each suggested revision is provided to the user.
To illustrate the problem, consider the following example.

\begin{figure}[htb]
\begin{center}
\includegraphics[scale=0.25]{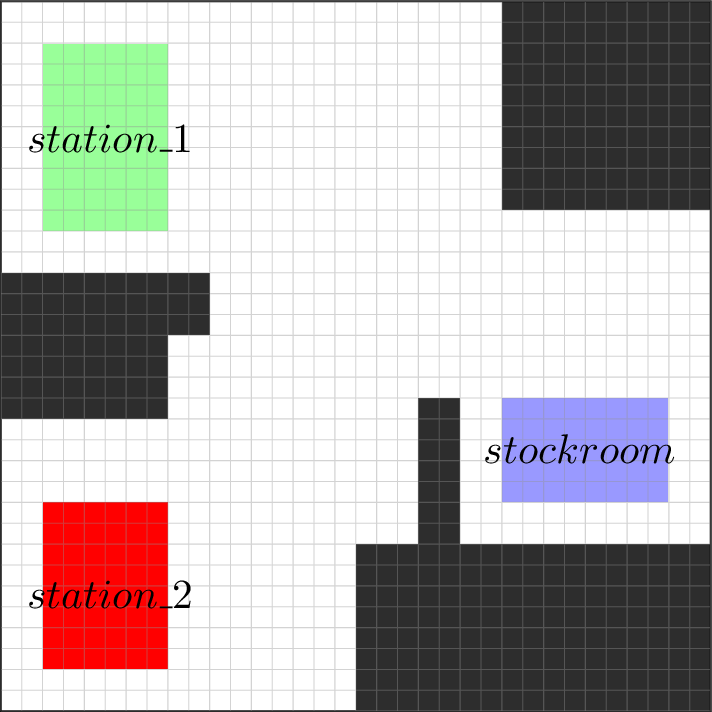}
\caption{Resupply example.}
\label{f:examplemap}
\end{center}
\end{figure}

\begin{example}
\label{ex:main_examp}
Consider a robot tasked with fetching parts in a factory setting, illustrated in Figure~\ref{f:examplemap}.  
Here, $stockroom$, $station\_1$ and $station\_2$ are regions in the workspace belonging to $\Y$ and $s1\_occupied$ and $s2\_occupied$ are sensors belonging to $\X$ indicating when the respective region is occupied.  The robot is required to visit the stockroom and the two workstations (system liveness) but avoid visiting those that are occupied (system safety).  If the robot is within a workstation, the environment is required to keep that station unoccupied (environment safety).  Also, the workstations are required to infinitely often be unoccupied (environment liveness).

The general specification $\varphi$ is as follows:
\begin{IEEEeqnarray*}{l,r}
\GG\FF stockroom \wedge \GG\FF station\_1 \wedge \GG\FF station\_2 \qquad & \triangleleft\textrm{ sys liveness}\\
\GG\FF \neg s1\_occupied \wedge \GG\FF \neg s2\_occupied \qquad
& \triangleleft\textrm{ env liveness}\\
\GG(s1\_occupied \implies\XX \neg station\_1)
& \triangleleft\textrm{ sys safety}\\
\GG(s2\_occupied \implies\XX \neg station\_2)
& \triangleleft\textrm{ sys safety}\\
\GG(station\_1 \implies \neg s1\_occupied) 
& \triangleleft\textrm{ env safety}\\
\GG(station\_2 \implies \neg s2\_occupied) 
& \triangleleft\textrm{ env safety}
\end{IEEEeqnarray*}
The specification $\varphi$ is realizable.  Suppose now we are given a fully-actuated planar robot governed by inertia, described by the system
\begin{IEEEeqnarray}{c't'c}
\ddot x = u && \ddot y = v,
\end{IEEEeqnarray}
where $(u, v) \in U$ are robot commands and $(x, y) \in \reals^2$ are the Cartesian robot coordinates.  We derive an abstraction $S_a$ of these dynamics in the configuration space $(x, y, \dot x, \dot y) \in W$ and obtain the formula $\varphi^{abs}$.  With $S_a$, we want to synthesize a realizable controller for $\varphi^{abs}$ that satisfies the task given an abstraction of these dynamics, with additional restrictions placed on the behaviors of the environment and the system.
\end{example}

%\begin{figure}[htb]
%\begin{center}
%\input{Figures/map}
%\caption{Resupply example.}
%\label{f:examplemap}
%\end{center}
%\end{figure}

Unrealizability can arise from a number of possible causes.  One possibility is that the robot's inertia may prevent it from avoiding collisions with either one of the workstations upon sensing that it is occupied.  Consider the case where $s1\_occupied$ is \false and the robot is moving toward $station\_1$.  Suppose that, when moving, the robot must pass through two grid cells to decelerate to a stop.  Then, the specification is unrealizable if $s1\_occupied$ is allowed to become \true when the robot is within two grid cells of $station\_1$.  This is an example of a {\em deadlock} behavior; the environment can force the system into certain states that have no legal transitions.

Another possibility is that the environment may toggle between states infinitely often, preventing the physical robot from making progress toward its goals.  For instance, suppose the robot is approaching $station\_1$ or $station\_2$ and the environment toggles the values of $s1\_occupied$ and $s2\_occupied$ infinitely often.  The robot in this case will be always changing directions, but unable to reach any of its goals before the environment changes once again.  This is an example of a {\em livelock} behavior; the robot is prevented from reaching its goals as a result of repeating sequence of environment inputs.  The behavior does not exist in the general formula because the topology graph always allows the robot to either remain in place or transit to an adjacent region once the environment has moved.

\section{Related Work} \label{sec:related}
We present a method that finds appropriate revisions to a specification when the inclusion of discrete abstractions cause unrealizability of specifications.  Our work intersects with two lines of research: synthesis with dynamics, and assumption mining for reactive synthesis.  We address related work in both of these areas below.

{\bf Synthesis with Dynamics.} \cite{TomLyg2000} were among the first to generate verified controllers preserving safety and reachability specifications for nonlinear dynamical systems.  In their approach, controllers are constructed by posing the problem as a differential game in hybrid systems (systems with mixed continuous and discrete states) analysis.  The work of~\cite{MBT05} made such ``reach-avoid'' problems tractable through viscosity solutions to time-dependent partial differential equations.

Progress has been made more recently in developing techniques for synthesis of controllers satisfying specifications with more expressive requirements than reachability and safety.  Namely,~\cite{KGFP_TRO09,KGWT11} introduce algorithms for reactive synthesis that take advantage of the {\em bisimulation property} allowing high-level controllers to be synthesized separately from the robot-specific low-level controllers.  In order to bridge the gap between the high- and low-levels of abstraction, tools for automatically synthesizing controllers in the continuous domain based on high-level specifications have been introduced recently in~\cite{fainekos2006,DKG_IJRR14}.  \cite{bhatia2010,maly2013} have solved this problem using multi-layered synthesis approaches, where certain parts of the control strategy are left open for an online planner to complete at runtime.  

Another perspective has been to incorporate the dynamics directly into the reactive synthesis process.  Rather than having an additional step for synthesizing continuous controllers that adhere to the behaviors of high-level controller,~\cite{liu2013} introduce a discrete abstraction of the underlying dynamics (\cite{girard2010,Rei11,ZPMT12}) directly into the synthesis process.  The problem of synthesizing controllers using abstractions of physical systems is well-studied: in provably-correct mission planning, researchers have considered robot dynamics ranging from simple single or double integrators (\cite{FGKGP09,KGFP_TRO09}) and piecewise linear models (\cite{tumova2010,YTCBB12}) to nonlinear (\cite{girard2010,bhatia2010,Wongpiromsarn2010,ZPMT12,wolff2013}), switched (\cite{liu2013}) and hybrid systems (\cite{maly2013}).  In~\cite{Kloetzer2008}, non-deterministic abstractions are used to synthesize non-reactive LTL formulas using model checking methods.  LTL synthesis for switched systems was considered in~\cite{liu2013,liu14}, where the authors propose methods for computing fine-grained abstractions and switching protocol synthesis for reactive tasks.  Our work is inspired by the nonlinear synthesis methods of~\cite{PGT08,ZPMT12,liu2013}.  Where our approach differs is that our synthesis algorithm takes specifications written agnostic to the robot and produces controllers that are applicable to a given robot platform, given its discrete abstraction.  We furthermore explore the case when the dynamics contribute to the unrealizability of such specifications and supply revisions in such cases.

{\bf Assumption Mining.}  Unrealizable specifications that are complicated to parse greatly benefit from automated tools for computing revisions that restrict the environment and system behaviors.  Our approach for computing such revisions is closely related to recent methods described in~\cite{fainekos11},~\cite{LiDS11}, and~\cite{AlurMT13}.  In~\cite{fainekos11}, a method is devised for determining the cause of unrealizability for non-reactive tasks and providing specification recommendations to the user.  In the reactive setting,~\cite{LiDS11} present a debugging method for unrealizable specifications based on templates (LTL formulas) mined from an environment counterstrategy.  In essence, a counterstrategy captures the possible behaviors for the environment for which there are no safe system moves that allow it to fulfill its goals.  The method in~\cite{AlurMT13} generates templates of the form ``$\FF$'' and ``$\FF\GG$'' automatically from the counterstrategy, yielding additional safety and liveness environment statements that remove all execution traces of the counterstrategy.  The work of~\cite{LiSSS14} apply the counterstrategy-based environment assumption mining technique to an early warning system in human-in-the-loop control systems, demonstrated in an autonomous driving scenario.  By removing the behaviors present in the counterstrategy, the modified environment is restricted in such a way as to permit the system to realize its goals under the strengthened assumptions, but can sometimes lead to specifications that no longer match the user's intent.  

Finding specific portions of the counterstrategy that lead to unrealizability is in general difficult without additional information or without post-processing the counterstrategy.  \cite{raman13} developed algorithms to extract cores (minimal LTL formulas) for explaining unrealizable specifications.  Similar to their work, we aim to identify two types of counterstrategy behaviors: deadlock and livelock.  Where the authors use an ``unrolling depth'' to discover counterstrategy states leading to livelock (a cycle of environment inputs locking the robot away from its goals), we apply a similar notion to finding states leading to deadlock (a counterstrategy state where no further system transitions exist).

Our approach differs from existing works in several other ways.  First, we adopt different strategy than~\cite{LiDS11,AlurMT13} for computing revisions, that depends on whether deadlock or livelock is being removed from the counterstrategy.  Second, we devise a means for efficiently computing revisions that reduces the number of times a counterstrategy needs to be synthesized.  Lastly, we explain the revisions in a simple-to-understand manner.  For example, the user will be provided with feedback in a structured format: ``The specification is realizable as long as a person is not present when the robot is within 2 meters of the Hallway.''

\section{Reactive Synthesis Under Nondeterministic Robot Abstractions} \label{sec:synth}
We synthesize reactive controllers by solving a two player game carried out between the environment (player 1) and the system (player 2) using the fixed-point algorithm described in~\cite{BJPPS12}.  As described in this section, we make modifications to this algorithm in order to accommodate the adversarial nature of the discrete abstraction and preserve a mapping between $\varphi$ and $\varphi^{abs}$.  We also distinguish the expressive properties of the proposed approach with alternative approaches (e.g.~\cite{liu2013}).  
%As described in this section, our aim is to perform synthesis on robot-specific formulas derived from general formulas written on a topology representation of the workspace.  Translating general formulas to robot-specific formulas necessitates that each statement in $\varphi$ has a valid, semantically-equivalent counterpart statement in $\varphi^{abs}$.  The synthesis approach in~\cite{liu2013} offers no mechanisms for semantic equivalence.
%For example, ``If you see a person, stay there'' requires the robot to stay in the current region as soon as it senses a person.  
%In this section, we formally define a new synthesis strategy for which a mapping exists between $\varphi$ and $\varphi^{abs}$, highlighting differences in the expressive properties over existing methods.

\begin{definition}[Controller Strategies for Non-deterministic Discrete Abstractions]
A controller in our abstractions-based paradigm is a non-deterministic finite-state machine $\A_{S_a} = (S,S_0,\X,\Y_a,\U_a,\delta,\gamma_\X,\gamma_\Y,\gamma_\U)$, where $S$, $S_0$, $\X$, $\Y_a$, $\U_a$, and $\gamma_\X$ are as described in Definition~\ref{d:strategy}, $\delta : S \times 2^{\X} \to 2^S$ is a state transition relation providing the set of possible states at the next position in the sequence given the current controller state and the current value of the environment input, $\gamma_\Y : S \to 2^{\Y_a}$ is a labelling function mapping controller states to a set of possible robot configurations evaluating to \true for transitions into that state, and $\gamma_\U : S \to 2^{\U_a}$ is a labelling function mapping controller states to a set of possible commands in that state.  
\end{definition}
Note that each state captures the possibility of non-determinism in the abstraction, obviating the sets-of-sets definition for $\gamma_\Y$.  The robot configuration may be regarded as being an additional environment input, but differs in that the value it takes is not defined until the system chooses a control input.  We therefore define an execution as a sequence of moves made by the environment and system, where the environment gets an extra move once the system has moved; in particular $\sigma = (\gamma_\X(s_0),\gamma_\U(s_0),\gamma_\Y(s_0))(\gamma_\X(s_1),\gamma_\U(s_1),\gamma_\Y(s_1)),\ldots$, where the robot motion at the current position in the sequence is a result of the environment input and robot command at the same position.  

The three-move formulation is in contrast to the two-move formulation of~\cite{liu2013}, where the state of the dynamical system at the current position in the sequence depends on the environment input and command at the {\em previous} position.  Recall that, as we require a controller $\A_{S_a}$ given the general-purpose formula $\varphi^e \implies (\varphi^s \wedge \varphi_t^{top})$, the three-move formulation allows for there to be a mapping between the subformulas $\varphi^e$, $\varphi^s$ and $\hat\varphi^e$, $\hat\varphi^s$.  For example, the safety guarantee $\varphi_t^s = \GG((person \wedge \XX\neg person) \implies \XX\neg r_1)$ has no equivalent counterpart $\hat\varphi_t^s$ in the method of~\cite{liu2013}.  The reason is that both $person$ and $r_1$ are environment variables, and the fact that the system is required to move {\em before} $r_1$ is decided forces $r_1$ to be delayed to the next position in the execution.  The translation is therefore $\GG((person \wedge \XX\neg person) \implies \XX\XX\neg r_1)$, which is not in GR(1).  Our strategy, on the other hand, allows for the environment to take an additional move after the system has moved, thereby preserving the mapping without delaying choosing $r_1$ to the next step.

We are able to assert that $\varphi^{abs}$ is realizable if, for all positions in the execution and for all possible valuations of the environment variables $\X$, there exists a command $\U_a$ such that all executions of $\A_{S_a}$ satisfy the specification for all $\Y_a$.  $\varphi^{abs}$ is unrealizable if there exists some environment behavior(s) such that no $\U_a$ can be found for which all robot configurations yield executions that satisfy the specification.

In order to determine if $\varphi^{abs}$ is realizable, our focus on the GR(1) (generalized reactivity (1)) fragment lends to efficient solutions, as described in~\cite{BJPPS12}.  Solving the GR(1) game is carried out by first introducing a {\em game structure}, defined as follows.

\begin{definition}[Game Structure]
A {\em game structure} is represented by the tuple $\G = (\V,\theta,\rho_e,\rho_r,\rho_s,\varphi_{win})$, where:
\begin{itemize}
\item $\V = \X\cup\U_a\cup\Y_a$ is the set of proposition valuations representing the position in the game;
\item $\theta\subseteq 2^\V$ is the set of initial positions;
\item $\rho_e \subseteq 2^{\V} \times 2^{\XX\X}$ is a transition relation defining the set of environment inputs allowed by $\varphi_t^e$ at the next position given the proposition values at the current position in the game;
\item $\rho_r \subseteq 2^{\V} \times 2^{\XX\U_a} \times 2^{\XX\Y_a}$ is a transition relation derived from the robot formula $\varphi_t^a$ defining the set of allowed robot configurations at the next position given the command at the next position and the robot configuration at the current position in the game;
\item $\rho_s \subseteq 2^{\V} \times 2^{\XX\V}$ is a transition relation defining the set of commands and robot configurations allowed by $\varphi_t^s$ at the next position given the proposition values at the current position in the game; and
\item $\varphi_{win}$ is the winning condition.
\end{itemize}
\end{definition}

The algorithm in~\cite{BJPPS12} proceeds by considering winning positions that satisfy the system and environment safety transition relations while leading the system toward its goals.  With the robot abstraction, we are faced with the additional step of ensuring that the positions chosen are safe for all possible robot configurations.  We therefore begin by specifying the following enforceable predecessor operator $\mathsf{Pre}$:
\begin{IEEEeqnarray*}{rCl}
\llbracket\mathsf{Pre} \W\rrbracket &:=& \{v \in 2^{\V} \mid \forall v_x \subseteq \X, \exists v_u \subseteq \U_a, \forall v_y \subseteq \Y_a : \\
&& ((v,v_x) \in \rho_e) \implies \left[ ((v,v_u,v_y) \in \rho_r) \implies \right. \\
&& \left. \left[ ((v,v_x,v_u,v_y) \in \rho_s) \wedge ((v_x, v_u, v_y) \in \llbracket\W\rrbracket) \right] \right]\}.
\end{IEEEeqnarray*}
%\JDC{separate the $\forall$ $\exists$ into appropriate parts of the formula?}
where $\llbracket\varphi\rrbracket$ denotes the set of positions for which the formula $\varphi$ evaluates to \true.
Intuitively, the enforceable predecessor is a set of positions enforcing that, for all environment inputs satisfying the environment transition relation, there exists a robot command such that all robot configurations bound to the transitions of the abstraction yields behaviors that remain safe, as long as the successor positions are taken from the set $\W$.  

We next define a set of winning positions $\V_{win}$ based on the $\mu$-calculus formula~\cite{BJPPS12} $\V_{win} = \nu \W_1. \bigwedge_{i_s\in I_s} \mu \W_2. \bigvee_{i_e\in I_e} \nu \W_3. N_{i_si_e}$, where $N_{i_si_e} = (B_{i_s}^s \wedge \mathsf{Pre}\W_1) \vee \mathsf{Pre}\W_2 \vee (\neg B_{i_e}^e \wedge \mathsf{Pre}\W_3)$ and $\nu$ and $\mu$ represent, respectively, the greatest and least fixpoint operators.  This formula ensures that there is a move that places the system strictly closer to one of its goals or one in which the system prevents one of the environment goals.  Note that the control strategy we derive must be consistent with the physical system.  The strategy $\A_{S_a}$ that we extract from the winning set of positions is therefore preventing from falsifying the discrete abstraction $\varphi_t^a$ (for example, those in which the robot is commanded to move to a configuration beyond $W$) by choosing only those control actions that also satisfy $\varphi_t^a$.

\subsection{Counterstrategies for Nondeterministic Robot Abstractions}
\label{s:counterstrategies}
When a specification is unrealizable, one may synthesize a counterstrategy to find a sequence of environment inputs and robot configurations that prevent the robot from fulfilling its specification.  
\begin{definition}[Environment Counterstrategies for Non-deterministic Discrete Abstractions]
We define an {\em environment counterstrategy} as a finite-state machine $\A^c = (Q^c,Q^c_0,\X,\Y_a,\U_a,\delta^c,\gamma^c_\U,\gamma^c_\X,\gamma^c_\Y)$, where 
\begin{itemize}
\item $Q^c$ is the set of counterstrategy states;
\item $Q_0^c\subseteq Q^c$ is the set of initial counterstrategy states;
\item $\X$, $\Y_a$ and $\U_a$ are sets of propositions in $AP$;
\item $\delta^c : Q^c \to 2^{Q^c}$ is a nondeterministic transition relation returning the set of counterstrategy states at the next position in the sequence given the current state;
%\item $\delta^c_e : Q^c \to 2^{\X}$ is a deterministic transition relation returning an environment input $\X$ at the next position in the sequence given the current state;
\item $\gamma^c_\U : Q^c \to 2^{\U_a}$ is a labelling function mapping counterstrategy states to the set of locomotion command propositions in $\U_a$ evaluating to \true for all transitions into that state;
\item $\gamma^c_\X : Q^c \to 2^{\X}$ is a labelling function mapping counterstrategy states to the set of environment propositions in $\X$ evaluating to \true in that state, and;
\item $\gamma^c_\Y : Q^c \to 2^{\Y_a}$ is a labelling function mapping counterstrategy states to the set of robot configurations in $\Y_a$ evaluating to \true in that state.
\end{itemize}
\end{definition}
We furthermore define $\delta^{c^{-1}} : 2^{Q^c} \to Q^c$ as the inverse transition relation mapping counterstrategy states to a set of predecessors; i.e. $\delta^{c^{-1}}(q') = \{q\in Q^c \mid q' \in \delta^c(q)\}$.

Our approach to synthesizing counterstrategies resembles that of~\cite{KonighoferHB09}, where we use a fixed point computation to determine realizability of the specification, then extract a strategy $\A^c$ from the set of positions that are winning for player 1.  We define the enforceable predecessor operator for the counterstrategy $\mathsf{Pre}^c$ as follows:
\begin{IEEEeqnarray*}{rCl}
\mkern-8mu \llbracket\mathsf{Pre}^c \W\rrbracket &:=& \{v \in 2^{\V^c} \mid \exists v_x \subseteq \X, \forall v_u \subseteq \U_a, \exists v_y \subseteq \Y_a : \\
&& ((v,v_x) \in \rho_e) \wedge \left[\left[ ((v,v_u,v_y) \in \rho_r) \wedge \right.\right. \\
&& \left. \left. ((v,v_x,v_u,v_y) \in \rho_s)\right] \implies ((v_x, v_u, v_y) \in \llbracket\W\rrbracket) \right] \}.
\end{IEEEeqnarray*}
The set of winning positions $\V^c_{win} = \X\cup\U_a\cup\Y_a$ for player 1 may then be computed from the formula $\V^c_{win} = \mu \W_1. \bigvee_{i_s\in I_s} \nu \W_2. \bigwedge_{i_e\in I_e} \mu \W_3. N^c_{i_si_e}$, where
\[
N^c_{i_si_e} = (\neg B_{i_s}^s \vee \mathsf{Pre}^c\W_1) \wedge \mathsf{Pre}^c\W_2 \wedge (B_{i_e}^e \vee \mathsf{Pre}^c\W_3), 
\]
%\JDC{todo: something may not be right.  check..}
A counterstrategy is synthesized by storing the sets $N^c_{i_si_e}$ for each $i_s$ and $i_e$ at the last pass of the fixed point operation.  Starting at the initial conditions $\varphi_i^e$, $\varphi_i^s$, for each state we identify an index $i_s$ of the liveness guarantee that is currently being prevented by the counterstrategy.  At a particular counterstrategy state $q$, we determine a successor $q'$ similar to~\cite{KonighoferHB09} as follows.  First, given the position $v$ at state $q$, we fix an assignment of inputs $v_x'\in\X$.  We next determine, for this fixed assignment, the set of winning configurations $v_y'\in\Y_a$ that belong to $N^c_{i_si_e}$ for every $v_u'\in\U_a$.  A check is made to determine if a liveness assumption has been fulfilled at $q$; if it has, then a new liveness assumption $i_e$ is selected.  For notational convenience, we define the set of successor values winning for player 1 at a counterstrategy state $q$ as 
$M_\X(q) = \{v_x'\in\X \mid \forall v_u' \in \U_a, \exists v_y'\in\Y_a : (v,v_x',v_u',v_y') \in N^c_{i_s(q)i_e(q)}, v = \gamma^c_\X(q)\cup\gamma^c_\U(q)\cup\gamma^c_\Y(q)\}$.  This set contains the set of inputs for which there exist command-configuration combinations that are winning for player 1.

%Executions in a counterstrategy lead either to one or more states from which the robot has no safe transitions (deadlock) or one or more cycles of states preventing the robot from fulfilling its goals (livelock).  In the remainder of this section, we discuss our approach for both cases.

\section{Generating Revisions to Unrealizable Specifications} \label{sec:envSafetyAssump}
In this section, we formalize our solution strategy for Problem~\ref{pr:prob1}.  
In similar fashion to~\cite{LiDS11,AlurMT13}, we make use of environment counterstrategies in order to search for environment assumptions that render the specification realizable.  
Our approach is outlined as follows: (1) from the specification $\varphi^{abs}$, we synthesize counterstrategies; (2) for each counterstrategy found, we compute environment and system transition subformulas $\psi_t^e$ and $\psi_t^s$ that prevent transitions to states in the counterstrategy from which the robot has no safe transitions (deadlock); (3) if a counterstrategy is found that is free of deadlock states, we compute liveness assumptions $\psi_g^e$ that restrict transitions to cycles of states preventing the robot from fulfilling its goals (livelock).  We introduce the following example to illustrate the major concepts discussed in this section.

\begin{figure}[htb]
\begin{center}
\subfloat[]{
%\begin{minipage}[b]{0.45\linewidth}
\centering
\label{f:examp1map}
\includegraphics[scale=0.25]{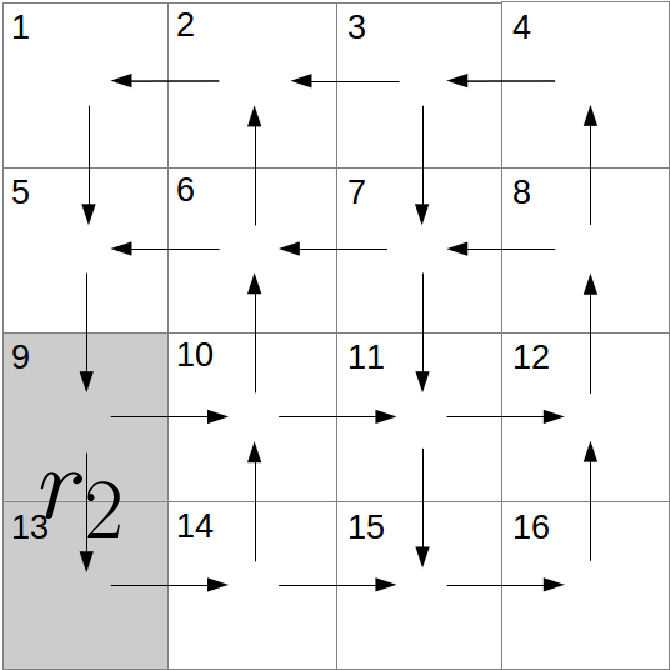}
%\end{minipage}
}
\\
\subfloat[]{
%\begin{minipage}[b]{0.45\linewidth}
\centering
\label{f:examp1aut}
\includegraphics[scale=0.25]{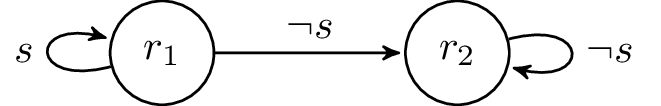}
%\end{minipage}
}
\caption{2-D example.  \protect\subref{f:examp1map} shows the workspace map and grid whose cells are labeled with the configuration variable.  The white grid cells denote $r_1$, while the gray denote $r_2$.  \protect\subref{f:examp1aut} shows the synthesized controller for $\varphi$.}
\end{center}
\end{figure}

\begin{example}
\label{ex:examp1}
Consider the workspace shown in Figure~\ref{f:examp1map}.  Given $\X = \{sen\}$ where $sen$ is the sensor input and $\Y = \{r1,r2\}$, we write a specification $\varphi$ requiring the robot to visit $r2$ (lower-left gray region) when $s$ is \false, but avoid $r2$ when $sen$ is \true.  Formally:
\begin{IEEEeqnarray*}{l,r}
%\GG\FF (\neg x0 \wedge \neg x1 \wedge y1) \wedge 
\GG\FF r2 & \triangleleft\ \varphi_g^s \\
\GG\FF \neg sen & \triangleleft\ \varphi_g^e \\
%\GG\FF (\neg s0) \wedge \GG\FF (\neg s1) \\
%\GG (s0 \implies \XX(\neg (\neg x0 \wedge \neg x1 \wedge y1))) \\
\GG (\XX sen \implies \XX\neg r2) \qquad\qquad & \triangleleft\ \varphi_t^s  \\
%\GG ((\neg x0 \wedge \neg x1 \wedge y1) \implies \XX\neg s0) \\
\GG (r2 \implies \XX\neg sen) & \triangleleft\ \varphi_t^e  \\
%\GG (s0 \implies \neg s1) \wedge \GG (s1 \implies \neg s0)
\true & \triangleleft\ \varphi_i^s \\
\true & \triangleleft\ \varphi_i^e
\end{IEEEeqnarray*}
The controller satisfying this specification is given in Figure~\ref{f:examp1aut}.

Given an abstraction where $\Y_a = \{x1,\ldots,x16\}$ is an encoding of the set of 2-D robot configurations, and $\U_a = \{N,S,E,W\}$ are the robot commands for motion in the four cardinal directions, we derive a new specification $\varphi^{abs}$, where: 
%\JDC{need to say something about the strengthening being trivial?}
\begin{IEEEeqnarray*}{l,r}
%\GG\FF (\neg x0 \wedge \neg x1 \wedge y1) \wedge 
\GG\FF (x9 \vee x13) & \triangleleft\ \hat\varphi_g^s \\
\GG\FF \neg sen & \triangleleft\ \varphi_g^e \\
%\GG\FF (\neg s0) \wedge \GG\FF (\neg s1) \\
%\GG (s0 \implies \XX(\neg (\neg x0 \wedge \neg x1 \wedge y1))) \\
\GG (\XX sen \implies \XX\neg (x9 \vee x13)) \qquad\qquad & \triangleleft\ \hat\varphi_t^s  \\
%\GG ((\neg x0 \wedge \neg x1 \wedge y1) \implies \XX\neg s0) \\
\GG ((x9 \vee x13) \implies \XX\neg sen) & \triangleleft\ \hat\varphi_t^e  \\
%\GG (s0 \implies \neg s1) \wedge \GG (s1 \implies \neg s0)
\true & \triangleleft\ \hat\varphi_i^s \\
\true & \triangleleft\ \hat\varphi_i^e
\end{IEEEeqnarray*}
The abstraction $\varphi_t^a$ appears as arrows in the figure.  
\end{example}

\subsection{Adding Revisions for Preventing Deadlock}
\label{s:deadlock}
%The general procedure works by finding deadlock conditions for an unrealizable specification: the specific conditions for which the environment can prevent the system from making further moves without violating system safety conditions.  
%Deadlock conditions are found using a counterstrategy, or a sequence of environment inputs that lead to unrealizability.  Using that counterstrategy to define a set of restrictive statements, then applying those statements to a set of configurations computed using the robot abstraction for which the same counterstrategy holds.

To prevent deadlock, we introduce a scheme to process a counterstrategy and extract a set of environment assumptions that remove deadlock behaviors.
%Note that, at deadlock, $\delta^c_s(q_{dead}^m,\delta^c_e(q_{dead}^m)) = \varnothing$; that is, there are no safe system moves from that state.  
Consider a counterstrategy $\A^c$ whose deadlock states are collected in $Q_{dead}$.  There exists an execution that eventually reaches a deadlock state $q_{dead}^i$; specifically when $\exists q' \in \delta^c(q_{dead}^i) : q' = \varnothing$.  We formally state this behavior as $\bigvee_{q^j\in\delta^{c^{-1}}(q_{dead}^i)}\FF\left(\psi_1(q^j) \wedge \XX\psi_2(q_{dead}^i)\right)$, where $\psi_1(q)$ and $\psi_2(q)$ are propositional representations for the subsets of positions at counterstrategy state $q$:
\begin{IEEEeqnarray*}{l,r}
\psi_1(q) = \mkern-10mu \bigwedge_{\pi\in \gamma^c_\U(q)\cup\gamma^c_\Y(q)} \mkern-10mu \pi \wedge \mkern-10mu \bigwedge_{\pi\in (\U_a\cup\Y_a)\backslash (\gamma^c_\U(q)\cup\gamma^c_\Y(q))} \mkern-10mu \neg\pi, \\
\psi_2(q) = \bigwedge_{\pi\in \gamma^c_\X(q)} \pi \wedge \bigwedge_{\pi\in \X\backslash\gamma^c_s(q)} \neg\pi.
\end{IEEEeqnarray*}
In words, $\phi_1(q)$ captures the command and configuration moves at state $q$, and $\psi_2(q)$ captures the environment input at $q$.
%\begin{IEEEeqnarray*}{l,r}
%\llbracket\psi^1(q)\rrbracket = \gamma^c_\U(q)\cup\gamma^c_\Y(q), \\
%\llbracket\psi^2(q)\rrbracket = \gamma^c_\X(q).
%\end{IEEEeqnarray*}

To remove the environment behaviors in the counterstrategy causing deadlock, we adopt the following formula:
\begin{IEEEeqnarray}{lr}
\bigwedge_{q^j\in\delta^{c^{-1}}(q_{dead}^i)} \mkern-10mu \GG \left(\psi_1(q^j) \implies \XX\neg\psi_2(q_{dead}^i)\right).
\label{e:envSafetySimple}
\end{IEEEeqnarray}
Before conjuncting each computed formula with $\psi_t^e$, a check is made to determine if it falsifies the left-hand side of $\varphi^{mod}$, i.e. there is no transition that satisfies $\hat\varphi^e \wedge \varphi_t^a \wedge \psi_t^e$.  If this is the case, the formula is discarded and it is not included as a conjunct in $\psi_t^e$.

\begin{figure}[htb]
\begin{center}
\subfloat[]{
\label{f:examp1CounterDeadlock}
\resizebox{0.6\columnwidth}{!}{
\includegraphics{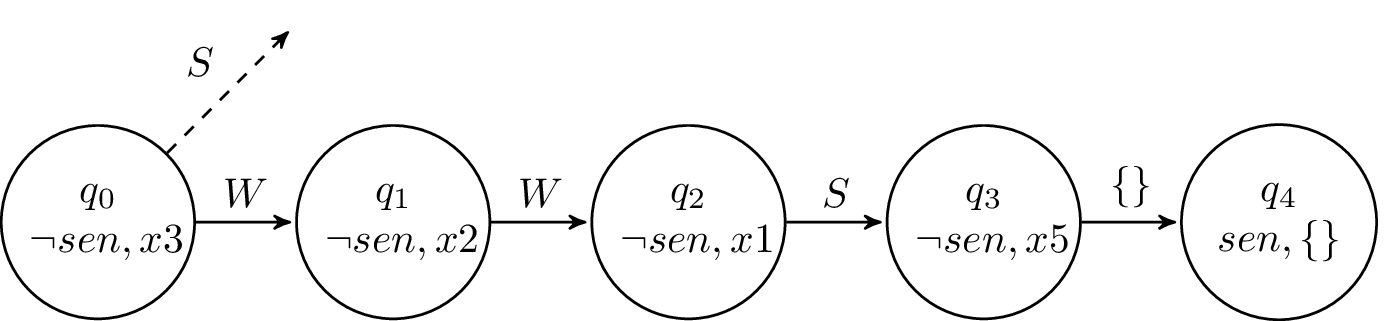}
}
}
\\
\subfloat[]{
\centering
\label{f:examp1CounterDeadMap}
\includegraphics[scale=0.25]{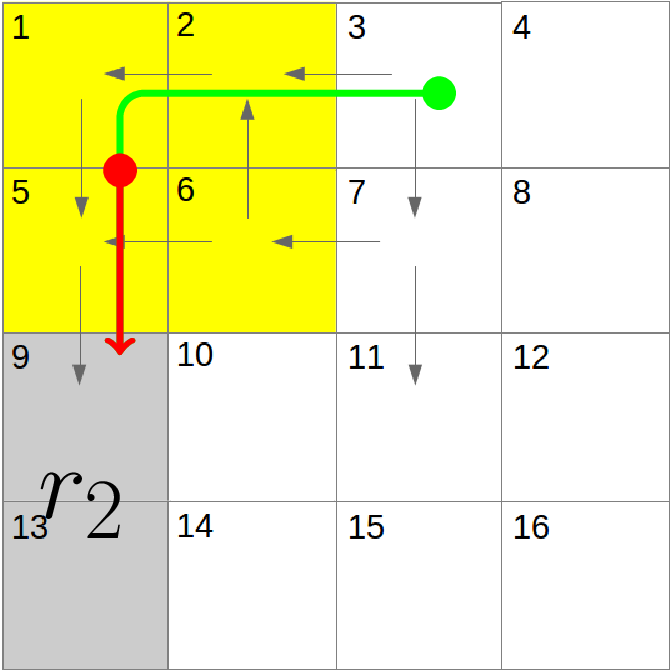}
}
\caption{\protect\subref{f:examp1CounterDeadlock} shows a partial counterstrategy for Example~\ref{ex:examp1deadlock} leading to deadlock; \protect\subref{f:examp1CounterDeadMap} shows a corresponding robot trajectory leading to deadlock, where the green part of the path denotes where $sen = \false$ and the red denotes where $sen = \true$.  The cells shaded yellow indicate configurations in which there are no sequence of commands that avoid reaching $r2$ eventually.}

\end{center}
\end{figure}

\begin{example}
\label{ex:examp1deadlock}
Returning to Example~\ref{ex:examp1}, suppose we obtain a counterstrategy containing the states $q_0,q_1,q_2,q_3,q_4$, as pictured in Figures~\ref{f:examp1CounterDeadlock} and~\ref{f:examp1CounterDeadMap}, starting in cell $x3$ with $sen = \false$.  One of the possible executions in this counterstrategy eventually leads the robot to cell $x4$ with the sensor $sen = \true$:
\begin{IEEEeqnarray*}{l,l}
\sigma = &(\{\varnothing\},\{\varnothing\},\{x3\}),(\{\varnothing\},\{W\},\{x2\}), (\{\varnothing\},\{W\},\{x1\}), \\
& (\{\varnothing\},\{S\},\{x5\}), (\{s\},\{\varnothing\},\{\varnothing\}).
\end{IEEEeqnarray*}
where the $i$th time step corresponds to $q_i$.  In this execution, the sensor $sen$ remains \false until the robot enters $x5$, at which point a transition in $\varphi_t^s$ is violated.  Hence $q_4$ is a deadlock state.  The formula $\FF(\XX sen \wedge \neg x1 \wedge S)$\footnote{We only make the \true action explicit ($S$ in this case), since mutual exclusion disallows the other actions from being activated at the same time.} is extracted by evaluating $\FF(\psi_1(q_4) \wedge \XX\psi_2(q_4))$.  The complement of this formula, $\GG( (\neg x1 \wedge S) \implies \XX\neg sen)$, is added as an additional environment assumption.  This assumption negates the behaviors in the counterstrategy for that particular deadlock state.  Being that there is only one deadlock state, we add no further assumptions.  Upon adding this revision to the environment assumptions, we determine that the modified specification is realizable.
\end{example}
%\JDC{note: we have to make sure to define an execution for the 'triple move' in the modified GR(1)}

Notice that the controller synthesized in Example~\ref{ex:examp1deadlock} produces executions that satisfy the specification but the system now assumes that the environment will always turn $sen$ \false whenever it reaches $x5$.  In the example, consider the behavior when the robot starts at $x7$ with $sen$ \true.  The execution of the robot in this case is 
\begin{IEEEeqnarray*}{l,l}
\sigma = &(\{sen\},\{\varnothing\},\{x7\}),(\{sen\},\{W\},\{x6\}), (\{sen\},\{W\},\{x5\}), \\
& (\{sen\},\{S\},\{x9\}), (\{\varnothing\},\{E\},\{x10\}),(\{\varnothing\},\{N\},\{x6\}),\ldots.
\end{IEEEeqnarray*}
In this execution, $sen$ remains \true and, as the robot moves toward $r2$, the environment eventually must set $sen$ to \false to be consistent with the added assumption.  When outside of $x5$, the robot follows the same sequence of moves regardless of the environment.  Note that the controller for the original, realizable specification $\varphi$ (Figure~\ref{f:examp1aut}) does not exhibit this behavior because there is no imposition on how the environment must behave based on the robot's configuration.  In that case, if $sen$ is \true, the robot waits in $r1$ until $sen$ becomes \false.

We remove such behaviors by disallowing a system transition whenever the same conditions found for the deadlock state hold at the state previous to deadlock.  The idea is to force the system to react conservatively to the newly-added environment revision as if it were a deadlock state.  Thus, when the robot is at a configuration previous to the deadlock state and the deadlock conditions hold currently, it will be forbidden from entering the configuration prior to deadlock.  On the other hand, when the environment is not currently producing the same conditions as at deadlock, $\psi_t^e$ forces the environment to ``play fair'' with the robot by not producing those conditions in the next step once the robot has made its move.

Formally, we disallow the behavior $\bigvee_{q^j\in\delta^{c^{-1}}(q_{dead}^i)} \FF(\XX\neg \psi_2(q_{dead}^i) \wedge \XX\neg \psi_1(q^j))$ by introducing an additional revision on $\varphi_t^s$  
\begin{IEEEeqnarray}{lCr}
\bigwedge_{q^j\in\delta^{c^{-1}}(q_{dead}^i)} \mkern-10mu &\GG & \left(\XX\neg\psi_2(q_{dead}^i) \implies \XX \psi_1(q^j) \right). \label{e:sysSafetySimple}
\end{IEEEeqnarray}
Such a revision places a safety restriction on the robot, preventing it from entering a {\em neighboring} state to a deadlock state whenever the environment is set to the same value for which deadlock occurs.  Doing this produces a specification that makes the system's behavior conservative; we are strengthening the conditions under which the robot may enter the neighboring state, when in fact the robot is not in any true danger of violating the original safety guarantees in $\varphi$ until it reaches $r_2$.  Nonetheless, if the specification is realizable, the system will be able to react to the environment as long as the actions/configurations are not included in those specified in $\bigwedge_{q^j\in\delta^{c^{-1}}(q_{dead}^i)} \psi_1(q^j)$.

If the modified formula is determined to be unrealizable and new deadlock states are found at a state $q^j\in\delta^{c^{-1}}(q_{dead}^i)$, then we once again return to the original set of circumstances specified in Problem~\ref{pr:prob1}.  We repeat the process in this section for as many times as required to eliminate deadlock states or when the specification is unrealizable.  We may, however, avoid repeated synthesis of counterstrategies by applying the assumption and guarantee revisions explained above to entire {\em subtraces} of a single counterstrategy (a a finite word of an execution trace for the counterstrategy).  To do this, we identify states for which there is no safe command to be taken such that there exists a subtrace that eventually visits states in $Q^c \backslash Q_{dead}$.  The search for deadlock revisions then reduces to a graph search on the counterstrategy, as summarized in Algorithm~\ref{a:extractCommit}.  The algorithm builds up a set of {\em deadlock-committed} states $Q_{commit}$ by adding, via a breadth-first search ($\mathsf{BFS}$ in line~\ref{line:bfs}), predecessor counterstrategy states from deadlock $Q_{dead}$ for which all locomotion commands lead to states in $Q_{commit}$.  For generating revisions and providing user feedback, we also maintain a mapping $Q_{reach} : Q_{commit} \to 2^{Q_{dead}}$ of deadlock states reachable from each $q\in Q_{commit}$.  The search continues until a fixed point of states is reached where no additional deadlock-committed states can be found, at which point $\mathsf{BFS}$ returns a tuple containing $Q_{commit}$ and $Q_{reach}$.  
%Note that such an action may not be part of the counterstrategy.  Therefore, our check on determining which predecessors are admissible is dependent on two criteria: 1) there is no action that leads to some other state in $\A_c$ not in $Q_{commit}$, or 2) there is an action allowed by the robot abstraction that is not in the counterstrategy.  
The precise condition under which the search terminates is when a $q\in Q^c$ is found such that:
\[
\exists q' \in \delta^c(q) : q' \notin Q_{commit}.
\]
Here, $Q_{commit}$ plays the role of $Q_{dead}$.  We therefore replace $Q_{dead}$ in the safety revisions~\eqref{e:envSafetySimple} and~\eqref{e:sysSafetySimple} with $Q_{commit}$.  To be precise, we replace~\eqref{e:envSafetySimple} with 
\begin{IEEEeqnarray}{l,r}
\bigwedge_{q^j\in Q_{commit}^i} \mkern-10mu \GG \left(\psi_1(q^j) \implies \mkern-22mu \bigvee_{q^k\in Q_{reach}(q_{commit}^i)} \mkern-22mu \XX\neg\psi_2(q^k)\right)
\label{e:envSafety}
\end{IEEEeqnarray}
and~\eqref{e:sysSafetySimple} with
\begin{IEEEeqnarray}{lCr}
\bigwedge_{q^j\in Q_{commit}^i} \mkern-10mu &\GG &\left( \mkern-0mu \bigwedge_{q^k\in Q_{reach}(q_{commit}^i)} \mkern-20mu \left( \XX\psi_2(q^k) \implies \XX \neg \psi_1(q^j) \right) \right), 
\label{e:sysSafety}
\end{IEEEeqnarray}
for each $q_{commit}^i \in Q_{commit}$.  Consider the example below.

\begin{algorithm}[!ht]
\caption{Computing deadlock-committed states.}
\label{a:extractCommit}
\begin{algorithmic}[5]
\Procedure{commitStates}{$Q_{dead}$}
\State Initialize $Q_{new},Q_{commit},Q_{reach}$ to $Q_{dead}$
\While {$Q_{new} \neq \varnothing$}
	\State $(Q_{new},Q_{reach}) \gets \mathsf{BFS}(\A_c, Q_{commit}, Q_{reach})$ \label{line:bfs}
	\State $Q_{commit} \gets Q_{commit} \cup Q_{new}$
\EndWhile
\State {\bf return} $Q_{commit}$, $Q_{reach}$
\EndProcedure
\end{algorithmic}
\end{algorithm}

\begin{figure}[htb]
\centering
\includegraphics[scale=0.25]{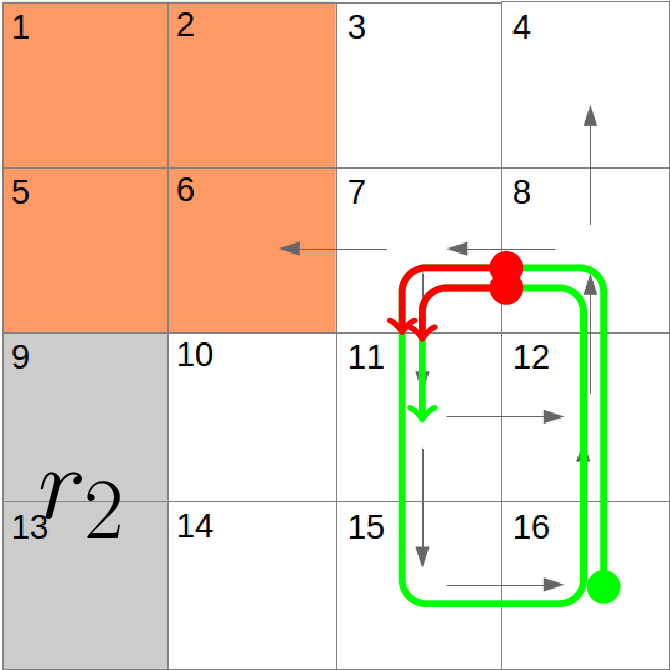}
\caption{Map showing configurations for which the revisions $\psi_t^e$ and $\psi_t^s$ from Example~\ref{ex:examp1sys} apply; a counterstrategy execution trace, as explained in Section~\ref{s:livelock}.  The green part of the path denotes where $sen = \false$ and the red denotes where $sen = \true$.}
\label{f:traces}
\end{figure}

\begin{example}
\label{ex:examp1sys}
Starting from the result of Example~\ref{ex:examp1deadlock}, we compute a set of four deadlock-commit states $Q_{commit} = \{q_1,q_2,q_3,q_4\}$ corresponding to the cells $\{x6,x2,x1,x5\}$.  We obtain the following $\psi_t^e$ formulas:
\begin{IEEEeqnarray}{l,r}
\GG((x5 \wedge S) \implies \XX\neg sen) \label{e:psite_first}\\
\GG((x1 \wedge S) \implies \XX\neg sen) \\ 
\GG((x2 \wedge W) \implies \XX\neg sen) \\
\GG((x6 \wedge N) \implies \XX\neg sen),
\end{IEEEeqnarray}
and the following $\psi_t^s$ formulas:
\begin{IEEEeqnarray}{l,r}
\GG(\XX sen \implies \XX\neg(x5 \wedge S)) \\
\GG(\XX sen \implies \XX\neg(x1 \wedge S)) \\
\GG(\XX sen \implies \XX\neg(x2 \wedge W)) \\
\GG(\XX sen \implies \XX\neg(x6 \wedge N)).\label{e:psits_last}
\end{IEEEeqnarray}
With these revisions added to $\psi_t^e$ and $\psi_t^s$ (highlighted orange in Figure~\ref{f:traces}, the modified specification eliminates the deadlock states present in the original counterstrategy.  Additionally, note that none of the revisions falsify the environment.
\end{example}

Upon synthesis, we find that a counterstrategy synthesized from this modified specification does not contain deadlock states.  In the next section, we discuss an approach to render the specification realizable by eliminating livelock behaviors.

\subsection{Adding revisions for Preventing Livelock}
\label{s:livelock}
In cases where the environment satisfies its own liveness conditions, the environment may be allowed to play in such a way that as the system cycles through an infinite sequence of states, the environment always keeps it away from one of its goals.  Consider the behavior of the robot when the above $\psi_t^e$ and $\psi_t^s$ formulas~\eqref{e:psite_first}--\eqref{e:psits_last} are introduced as revisions.  Starting at $x16$, the following behavior is possible:
\begin{IEEEeqnarray*}{l,l}
\sigma = &(\{\varnothing\},\{\varnothing\},\{x16\}),(\{\varnothing\},\{N\},\{x12\}), (\{\varnothing\},\{N\}, \{x8\}), \\
& (\{sen\},\{W\},\{x7\}), (\{\varnothing\},\{S\},\{x11\}),(\{\varnothing\},\{S\},\{x15\}), \\
& (\{\varnothing\},\{E\},\{x16\}),(\{\varnothing\},\{N\},\{x12\}),\ldots.
\end{IEEEeqnarray*}
In this execution(shown in Figure~\ref{f:traces}), the robot eventually cycles indefinitely between six cells in the workspace.  Whenever the robot visits the cell $x7$, the environment activates $sen$, forcing the robot to move $S$ to avoid violating the safety guarantee revision in~\eqref{e:psits_last}.  The environment is then able to satisfy its liveness goal ($\GG\FF(\neg sen)$), while preventing the robot from achieving its goal of reaching $r_2$.

Once we obtain a counterstrategy free of deadlock states, we generate environment assumptions that remove the counterstrategy executions that exhibit livelock.  The idea is to selectively choose states in the counterstrategy for which the robot still has winning actions to take and then apply liveness assumptions at those states to ensure that, always eventually, the robot is allowed to take these actions.  We employ the result of the innermost fixed point computation $M_\X(q)$ stored when solving the counterstrategy game to find any states for which there exists an assignment of environment inputs that are not winning for the environment.  Note that, for these inputs, there exists a command the robot may take which is winning for the system.  We then prevent the environment from always making such assignments at these counterstrategy states.

Formally, for all $q\in Q^c$, our goal is to find a subset $Q_{cut} \subseteq Q^c$ for which $v_x'\notin M_\X(q)$, where $M_\X(q)$ are the set of environment inputs at a state $q\in Q^c$ that are winning for player 1.
Consequently, for any $q \in Q_{cut}$ there are some $v_x'\subseteq\X$ and $v_u'\subseteq\U_a$ that lead to a position that is not winning for player 1 (i.e. $v_x'\notin M_\X(q)$).  One can think of $Q_{cut}$ as being those counterstrategy states where it is possible that the environment has been able to ``cut away'' a command that will allow the robot to proceed to its next goal by applying some environment input.

%As a final step, we remove from $M^c_{i_si_e}$ those actions for which a configuration $\XX\Y_a$ intersects with one of the 

Using $Q_{cut}$, we formulate a set of liveness assumptions that restrict the environment from {\em always} behaving in a manner that prevents the system's progress toward its goals.  Notice that $Q_{cut}$ contains all states for which there is an environment and system move not in player 1's strategy; however, not all such moves are necessarily winning for player 2.  For instance, a state in $Q_{cut}$ could yield an environment input that does not allow player 1 to move strictly closer to its goal yet only allow system moves that place the system further away from its goal.  

We therefore form a set $P_{cut} \subseteq Q_{cut}$ for which the robot has safe commands that are winning for the system.  We use $Q_{commit}$ (from the deadlock counterstrategy) to define the set of states where there exist system moves that lead the system closer to its goals.  We populate $P_{cut}$ as follows:
%\begin{IEEEeqnarray}{lCr}
%P_{cut} &=& \{q\in Q_{cut} \mid \exists v_a\in V_a, \exists q' \in Q_{commit} : \nonumber \\
%&& \delta_a(v_a,\gamma^c_\Y(q)) \cap \gamma^c_\Y(q') \neq \varnothing\}. \label{e:cut_states}
%\end{IEEEeqnarray}
\begin{IEEEeqnarray}{lCr}
P_{cut} &=& \{q\in Q_{cut} \mid \exists v_a\in V_a, \exists q' \in Q_{commit},
 \exists q_a \in \delta_a(v_a,\gamma^c_\Y(q)) : \nonumber \\
&& \forall \pi\in\Y_a, \pi \in \gamma_a^\Y(q_a) \textrm{ iff } \pi \in \gamma^c_\Y(q')\}. \label{e:cut_states}
\end{IEEEeqnarray}
We then apply the environment liveness assumption 
\begin{IEEEeqnarray}{l,r}
\label{e:envLiveness}
\GG\FF \bigvee_{q^i\in P_{cut}} \left( \psi_1(q^i) \wedge \mkern-20mu \bigwedge_{q^j\in\delta^c(q^i,\gamma^c_\Y(q^i))} \mkern-20mu \XX\neg \psi_2(q^j) \right).
\end{IEEEeqnarray}
This liveness formula disallows the environment from always behaving in a way that denies the system from taking action that lead it closer to its goals, when the robot is in a configuration where there is such an action to be taken.  %Note also that the existence of $P_{cut}$ on a counterstrategy having been already found for deadlock, which may not always be the case in general.

\begin{figure}[htb]
\centering
\includegraphics[scale=0.25]{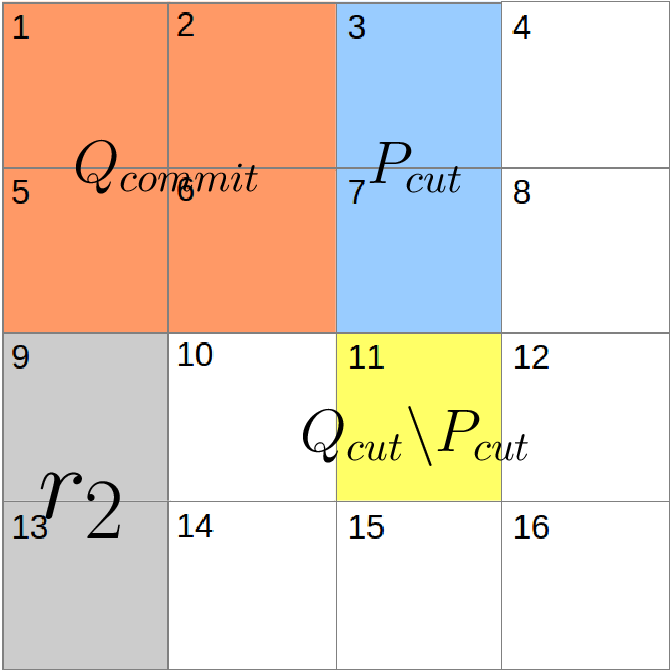}
\caption{Map showing regions associated with cut states from Example~\ref{ex:examp1liveness}.}
\label{f:cutregions}
\end{figure}

\begin{figure}[htb]
\begin{center}
\centering
\resizebox{0.6\columnwidth}{!}{
\includegraphics{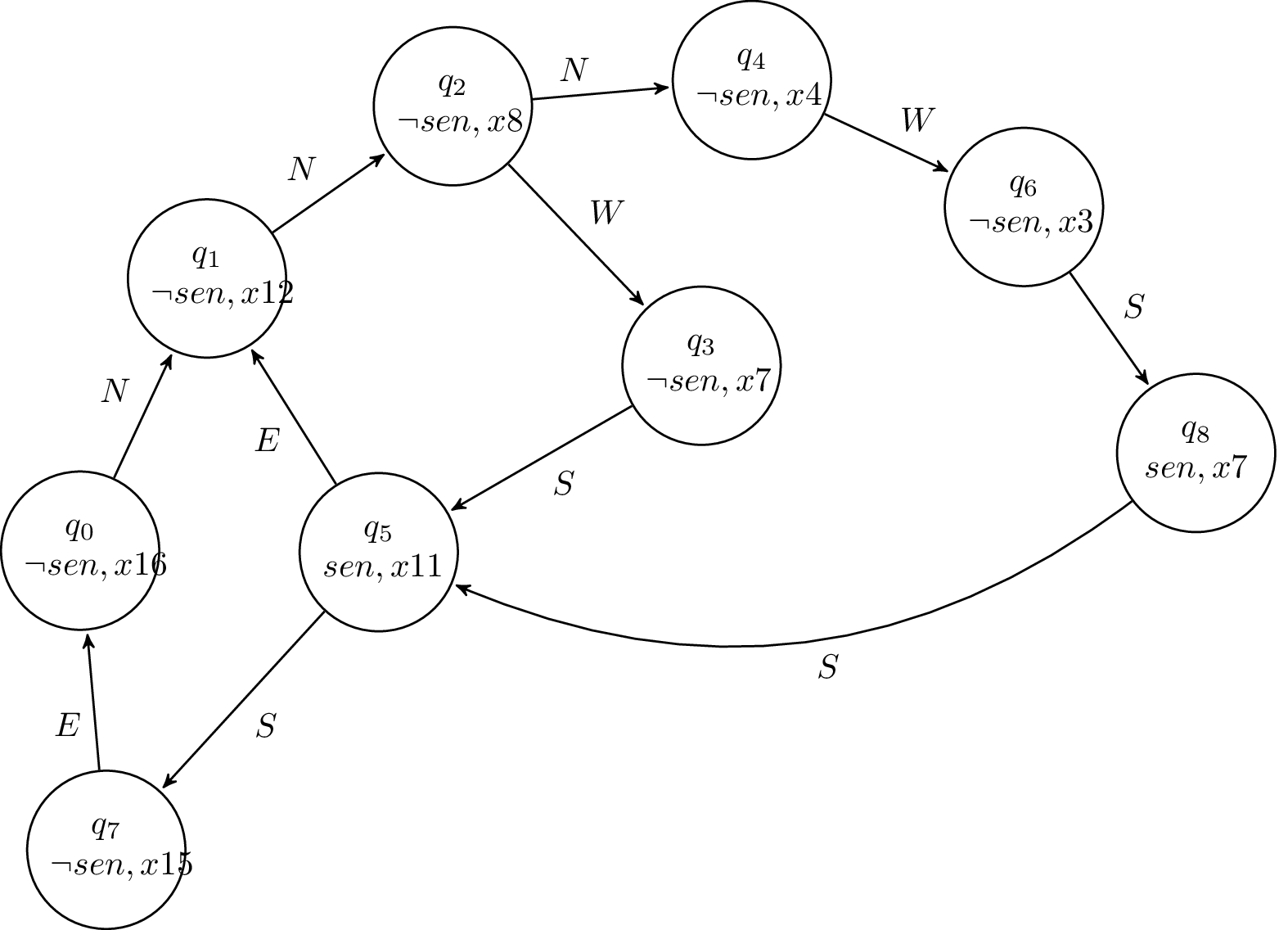}
}
\caption{Deadlock-free counterstrategy for Example~\ref{ex:examp1liveness}.}
\label{f:examp1CounterLivelock}
\end{center}
\end{figure}

\begin{example}
\label{ex:examp1liveness}
With the specification $\varphi^{abs}$ in Example~\ref{ex:examp1} along with the revision~\eqref{e:psite_first}--\eqref{e:psits_last}, a counterstrategy is extracted as pictured in Figure~\ref{f:examp1CounterLivelock}.  The set $Q_{cut}$ consists of four states, $\{q_3,q_5,q_6,q_8\}$.  Of these, $\{q_3,q_6,q_8\}$ are states in which the robot has an action (move $W$) taking the robot closer to $r_2$ that the environment can prevent.  These are found from Example~\ref{ex:examp1deadlock} and are collected in $P_{cut}$.  We also find state $\{q_5\}$, for which there is an environment move keeping it from immediately realizing an environment goal ($\GG\FF(\neg sen)$) but does not lead the system closer to its goal.  We apply environment liveness revisions $\psi_g^e$ to the set $P_{cut}$:
\begin{IEEEeqnarray}{lr}
\GG\FF &\left( (x7 \wedge W \wedge \XX\neg sen) 
\vee (x3 \wedge W \wedge \XX\neg sen) 
\vee (x7 \wedge S \wedge \XX\neg sen) \right) .
\end{IEEEeqnarray}
The three above formulas correspond to regions appearing as blue regions in Figure~\ref{f:cutregions}.  Adding this final revision produces a specification $\varphi^{mod}$ that is realizable. 
%\JDC{paragraph describing/showing the execution?}
\end{example}

\subsection{User Feedback}
\label{s:feedback}
Before modifying the specification with the computed deadlock revisions, we alert the user of the consequences of these revisions.  Given this information, the user may choose to accept these if they are, in fact, consistent with the original design intent.  We provide feedback in the form of statements such as ``Keep sensor $sen$ \false if the robot enters to within $N$ meters of $r2$'' instead of raw LTL formulas.

For the combined environment and system revisions for deadlock, we use the mapping between cells and regions from the robot abstraction to assist in forming this metric.  For each labeled workspace region $R_i\in\R$, we mark those deadlock states (if any exist) from whose predecessors there exists a robot command $v_a\in V_a$ (Definition~\ref{d:abstraction}) to reach $R_i$.  Those marked deadlock states are collected in the set $P_{dead}(R_i)$, defined formally as:
%\begin{IEEEeqnarray*}{lCr}
%P_{dead}(R_i) &=& \{q\in Q_{dead} \mid \forall q'\in\delta^{c^{-1}}(q), \exists v_a\in V_a : \\
%&& \delta_a(v_a,\gamma^c_\Y(q')) \cap \gamma_a(R_i) \neq \varnothing\}.
%\end{IEEEeqnarray*}
\begin{IEEEeqnarray*}{lCr}
P_{dead}(R_i) &=& \{q\in Q_{dead} \mid \forall q'\in\delta^{c^{-1}}(q), \exists v_a\in V_a, 
 \exists q_a' \in \delta_a(v_a,\gamma^c_\Y(q')) : \\
&& \forall \pi \in \Y_a, \pi \in \gamma_a^\Y(q_a') \textrm{ iff } \pi \in\gamma_a(R_i)\} .
\end{IEEEeqnarray*}  
%\JDC{TODO: $\gamma^\Y_a$ and $\gamma^c_\Y$ (exchange of sup's and sub's) might get annoying to read. Fix these!}

In the fixed point computation in Algorithm~\ref{a:extractCommit}, we keep track of each deadlock state reachable from each state added to $Q_{commit}$.  We use this stored information to find the distances between each robot configuration in $Q_{commit}$.  Rather than provide the user with a detailed set of conditions under which the environment would be required to adhere for the generated revisions to be satisfied, we advocate for simplicity by computing a conservative upper-bound on robot configurations where the environment restrictions must hold.
For each $R_i$ corresponding to a robot configuration that satisfies some deadlock state in $Q_{dead}$, we find the relative proximity to a deadlock condition (in terms of physical coordinates) by finding the maximal pairwise distance between any states affected by the deadlock revisions:
\begin{IEEEeqnarray}{lCr}
\label{e:furthestStates}
(q^\star_i, q_{dead,i}^\star) &=& \underset{\substack{q\in Q_{commit}, \\ q'\in Q_{reach}(q) \cap P_{dead}(R_i)}}{\arg\max} \vert \gamma^c_\Y(q) - \gamma^c_\Y(\delta^{c^{-1}}(q')) \vert .
\end{IEEEeqnarray}
Here, $\vert v_y \vert$ is the Euclidean norm of the real-valued abstraction state $q_a\in Q_a$ represented by a set of propositions $v_y\subseteq\Y_a$ that are \true in that state.  The pair of counterstrategy states $q^\star_i$ and $q_{dead,i}^\star$ are those corresponding to a revision for region $R_i$ where the distance is greatest, under the constraint that $q_{dead,i}^\star$ is a deadlock state that is reachable from $q^\star_i \in Q_{commit}$.  Note that the distance between the configurations of the two states is:
\begin{IEEEeqnarray*}{lCr}
%\label{e:distance}
dist_i &=& \vert \gamma^c_\Y(q^\star_i) - \gamma^c_\Y(\delta^{c^{-1}}(q_{dead,i}^\star)) \vert.
\end{IEEEeqnarray*}

%\JDC{the discussion below assumes that the counterstrategy always chooses same environment input for all deadlock states concerning the same region!}
The final step is to correlate each unique region $R_i$ to the environment proposition assignments prevented by the safety assumption revisions $\psi_t^e$.  Those prevented assignments are given in the formula $\psi_2(q_{dead,i}^\star)$.  That is, the added environment assumptions prevent the environment from triggering the combination $\psi_2(q_{dead,i}^\star)$.  The data provided to the user is represented by the triple $(R_i,dist_i,\psi_2(q_{dead,i}^\star))$.  The triple can be displayed to the user as follows: ``If the robot is within $dist_i$ of workspace region $R_i$, then the generated deadlock revisions (for a given counterstrategy) will be satisfied if the environment is not set to $\psi_2(q_{dead,i}^\star)$.''  Note that this metric supplies a sufficient but not necessary condition for satisfying the revisions.  That is, there might be executions where the system enters within $dist_i$ with any environment setting yet still be able to satisfy the revision formulas.

\begin{example}
\label{ex:feedback}
In the result of Example~\ref{ex:examp1sys}, let $q_{dead}$ be the deadlock state computed by the counterstrategy corresponding to the configuration $x9$, and designate $Q_{commit} = \{q_1,q_2,q_3,q_4\}$ as the set of commit states for this deadlock.  For workspace region $r_2$, $P_{dead}(r_2) = q_{dead}$, and $Q_{reach}(q_i) = q_{dead}$ for $i=1,\ldots,4$.  We next determine the pair $(q^\star_2, q_{dead,2}^\star)$ to be
\begin{IEEEeqnarray*}{lCr}
(q^\star_2, q_{dead,2}^\star) &=& \arg\max
\left\{\left| \begin{pmatrix}0\\1\end{pmatrix} - \begin{pmatrix}0\\2\end{pmatrix} \right|, 
\left| \begin{pmatrix}0\\0\end{pmatrix} - \begin{pmatrix}0\\2\end{pmatrix} \right|,
 \left| \begin{pmatrix}1\\0\end{pmatrix} - \begin{pmatrix}0\\2\end{pmatrix} \right|,
\left| \begin{pmatrix}1\\1\end{pmatrix} - \begin{pmatrix}0\\2\end{pmatrix} \right| \right\} 
 = (q_2, q_{dead}), 
\end{IEEEeqnarray*}
where the subscript 2 in the $^\star$ variables is used to signify the fact that the variables apply to region $r_2$.  Assuming $\eta = 1 m$, the corresponding distance is $dist_2 = \sqrt{1^2 + 2^2} ~= 2.2 m$.  Finally, reflective of the revisions in~\eqref{e:psite_first}--\eqref{e:psits_last}, we note the subformula $\psi_2(q_{dead,2}^\star) = sen$.

Therefore, the LTL formulas in~\eqref{e:psite_first}--\eqref{e:psits_last} are are summarized as: ``If the robot enters to within $2.2 m$ of $r_2$, never set environment variable $sen$ to \true.''
\end{example}

\subsection{Summary of the Approach}
The approach described in Sections~\ref{s:counterstrategies}--\ref{s:feedback} may be combined into an automatic approach for synthesis assisted by user feedback.  We outline this process in Algorithm~\ref{a:refine}.  The approach has three main stages: computing revisions for deadlocks, composing an explanation of the revisions, and computing revisions for livelocks.  Realizability is checked at every iteration of a while loop, terminating either when a specification is realizable or there are no further revisions to be computed (in this case, the counterstrategies from one iteration to the next will be identical).  Revisions for eliminating deadlock in the counterstrategy are computed by applying~\eqref{e:envSafety} and~\eqref{e:sysSafety} to the deadlock-committed states $Q_{commit}$.  If these revisions falsify the environment and system, they are removed.  The second step is to provide feedback to the user.  Depending on the user's response, the revisions are either applied or discarded.  

%\addedJD{
The third step in the approach is to generate revisions for liveness as computed in~\eqref{e:envLiveness}.  Once a candidate liveness assumption is computed, it is checked in Lines~\ref{line:envFalseChkSt}--\ref{line:envFalseChkEnd} to ensure that the system's strategy does not contain a sequence of moves that cause the new liveness condition to be falsified.  In such cases, $\mathsf{realizable}$ returns \false, and the candidate liveness is removed.  We refer the reader to Algorithm 5 of~\cite{RKG13} for further details.  The user may elect to accept or discard this formula.  %}  
Note that, if the specification is unrealizable and the counterstrategy is the same between iterations of the while loop, this means that no revisions have been found that meet the user's criteria or do not falsify the specification.  In this case, the algorithm terminates with an {\em unrealizable} output.

\begin{algorithm}[!ht]
\caption{Synthesizing revisions for an unrealizable specification $\varphi^{abs}$}
\label{a:refine}
\begin{algorithmic}[5]
\Procedure{synthRevisions}{$\varphi^{abs}$}
\State Initialize $\psi_t^e,\psi_t^s,\psi_g^e$ to \true
\State $(realiz,\A_c^1,M_\X) \gets \mathsf{ctrStrategy}(\varphi^{abs})$
\State $\A_c^0 \gets \false$, $m \leftarrow 0$ 

\While {$\neg realiz \wedge (\A_c^m \neq \A_c^{m-1})$}

\State \Comment \textit{Step 1: Eliminate deadlocks}
\State $Q_{commit} \gets \mathsf{commitStates(\A_c)}$
\ForAll {$q_{commit}^i \in Q_{commit}$ \label{line:forallCommit}}
	\State $\psi_{t,cand}^e, \psi_t^e \gets $ Eq.~\eqref{e:envSafety}
	\State $\psi_{t,cand}^s, \psi_t^s \gets $ Eq.~\eqref{e:sysSafety}
	\If {$\neg (\varphi^e \wedge \varphi_t^a \wedge \psi_g^e \wedge \psi_t^e \wedge \psi_{t,cand}^e)$ or $\neg (\hat\varphi^s \wedge \psi_t^s \wedge \psi_{t,cand}^s)$}
		\State $\psi_t^e \gets \psi_t^e \backslash \psi_{t,cand}^e$
		\State $\psi_t^s \gets \psi_t^s \backslash \psi_{t,cand}^s$
	\EndIf
\EndFor \label{line:forallCommitEnd}

\State \Comment \textit{Step 2: User feedback}
\ForAll {$R_i \in \R$ \label{line:forallR}}
	\State $(q^\star_i,q_{dead}^\star) \gets$ Eq.~\eqref{e:furthestStates} 
	\State $dist_i \gets \vert \gamma^c_\Y(q^\star) - \gamma^c_\Y(q_{dead}^\star) \vert$ 
	\State {\bf print} $(R_i,dist_i,\psi_2(q_{dead,i}^\star))$
\EndFor

\If {user accepts deadlock revisions}
	\State $\varphi^{mod} \gets$ Eq.~\eqref{e:varphimod}
	\State $(realiz,\A_c^m,M_\X) \gets \mathsf{ctrStrategy}(\varphi^{mod})$ 
\EndIf

\State \Comment \textit{Step 3: Eliminate livelocks}
\State $Q_{cut} \gets \{q \in Q^c \mid \exists v_x' \notin M_\X(q)\}$
\State $P_{cut} \gets $ Eq.~\eqref{e:cut_states}
\State $\psi_{g,cand}^e, \psi_g^e \gets $ Eq.~\eqref{e:envLiveness}
\If {$\neg(\varphi^e \wedge \varphi_t^a \wedge \psi_g^e \wedge \psi_t^e \wedge \psi_{g,cand}^e)$}
	\State $\psi_g^e \gets \psi_g^e \backslash \psi_{g,cand}^e$ 
\Else \label{line:envFalseChkSt}
	\State $\varphi^{try} \gets$ Eq.~\eqref{e:varphimod}
	\State $realiz \gets \mathsf{realizable}(\varphi^{try})$ 
	\If {$\neg realiz$}
		\State $\psi_g^e \gets \psi_g^e \backslash \psi_{g,cand}^e$ 
		\State \Comment \textit{System falsifies environment liveness}
	\EndIf
\EndIf \label{line:envFalseChkEnd}

\If {user accepts livelock revisions}
	\State $\varphi^{mod} \gets$ Eq.~\eqref{e:varphimod}
	\State $(realiz,\A_c^m,M_\X) \gets \mathsf{ctrStrategy}(\varphi^{mod})$ 
\EndIf
\State $m++$ 
\EndWhile
\State {\bf return} $\varphi^{mod}$ \label{line:returnBuff}  
\EndProcedure
\end{algorithmic}
\end{algorithm}

\section{Example} \label{sec:example}

In this section, we demonstrate the revision synthesis approach in an example scenario in which revisions to an unrealizable task specification are computed and added to the specification based on guidance provided by the user.
We generate abstractions using the Pessoa Toolbox.\footnote{https://sites.google.com/a/cyphylab.ee.ucla.edu/pessoa/}  For synthesis, we use the Slugs Synthesis Tool, part of the LTLMoP Toolkit;\footnote{https://github.com/LTLMoP/slugs} the code used in this paper may be found at https://github.com/jdc1177/slugs.

We return to the factory scenario in Example~\ref{ex:main_examp} using the workspace in Figure~\ref{f:examplemap}.  To carry out this task, we select a robot described by a unicycle model that is governed by the kinematic relationship:
\begin{IEEEeqnarray*}{c't'c't'c}
\label{eq:uniModel}
\dot x = v\cos\theta, && \dot y = v\sin\theta, && \dot\theta = \omega,
\end{IEEEeqnarray*}
where the $x$ and $y$ are the Cartesian displacements in meters, $\theta$ is the orientation angle, and $v$ and $\omega$ are, respectively, the forward and angular velocity inputs to the system.  The car model is subjected to the constraint where it may only move with {\em positive} forward velocity (it cannot stop).  An abstraction is generated for the three-dimensional configuration space and two-dimensional input space consisting of $2.2\times 10^6$ states, with the chosen values $\eta = 0.15$, $\mu = 0.2$, $\tau = 0.35$.

\begin{figure}[htb]
\centering
\resizebox{0.6\columnwidth}{!}{
\includegraphics{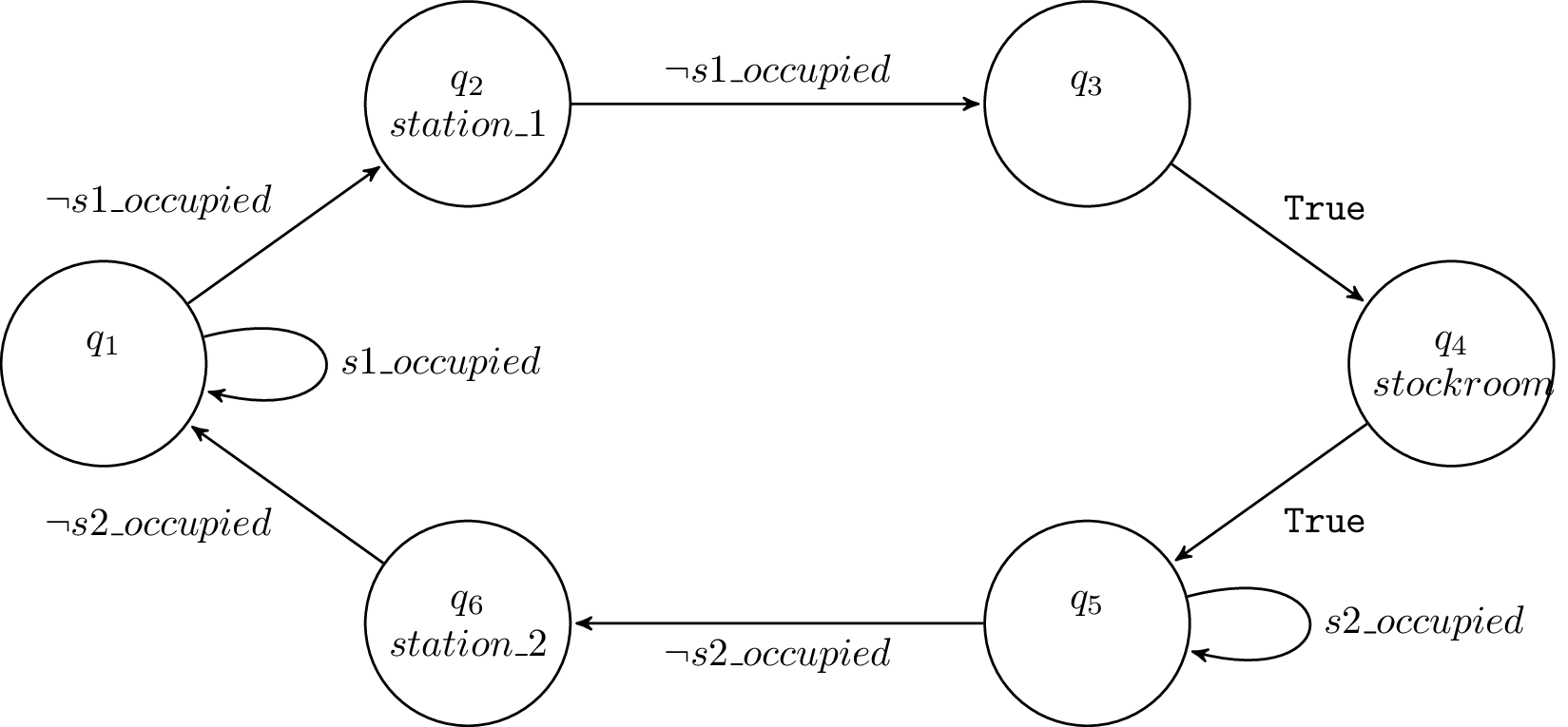}
}
\caption{Controller for $\varphi$ in Example~\ref{ex:main_examp}. Edges are labeled with the disjunction of assignments in $\X$ that may be assumed for that transition.}
\label{f:aut_main_examp}
\end{figure}

\begin{figure*}[htb]
\centering
\subfloat[]{
\label{f:traj1}
\centering
\includegraphics[scale=0.15]{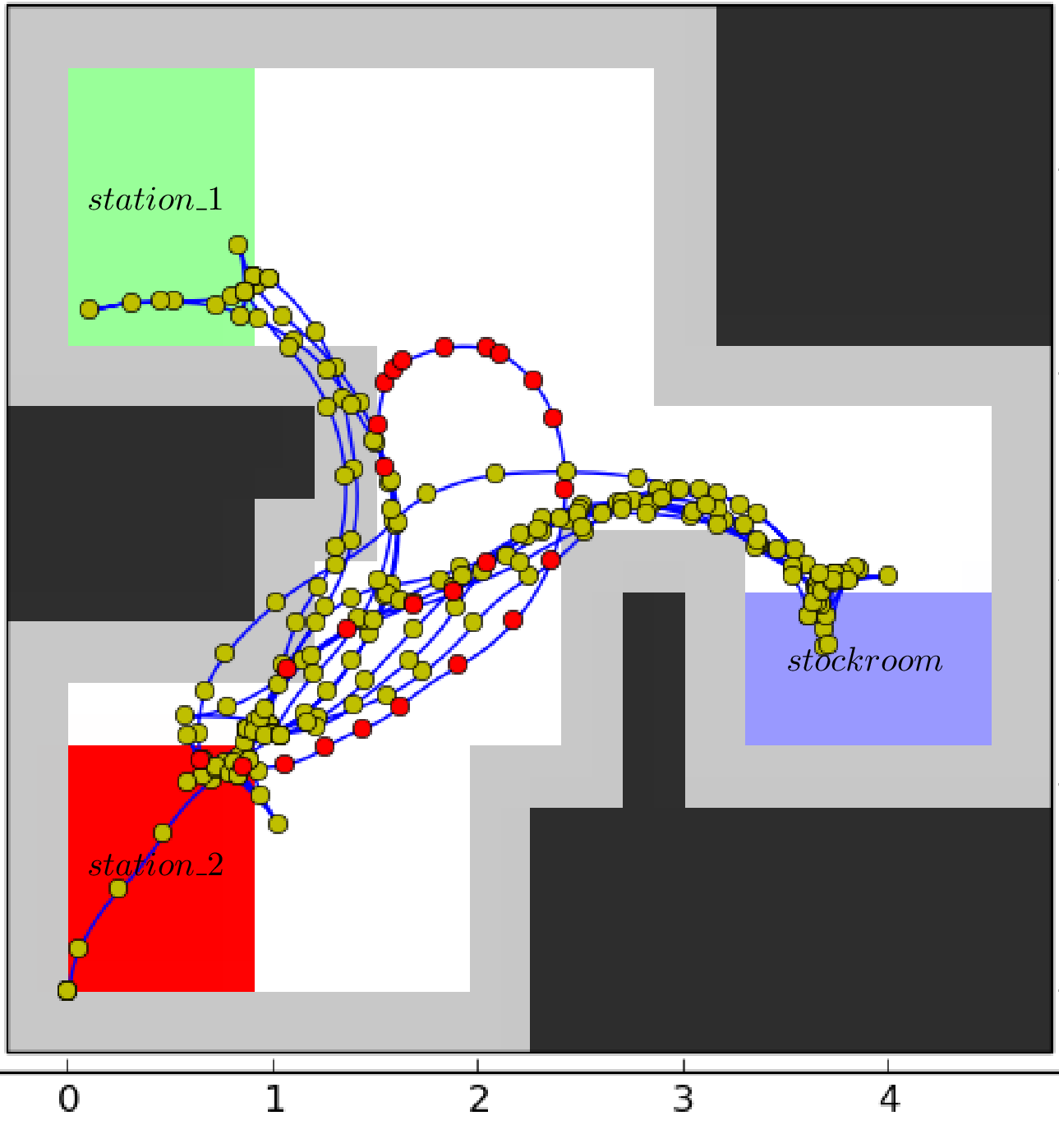}
}
\hspace{1cm}
\subfloat[]{
\label{f:traj2}
\centering
\includegraphics[scale=0.15]{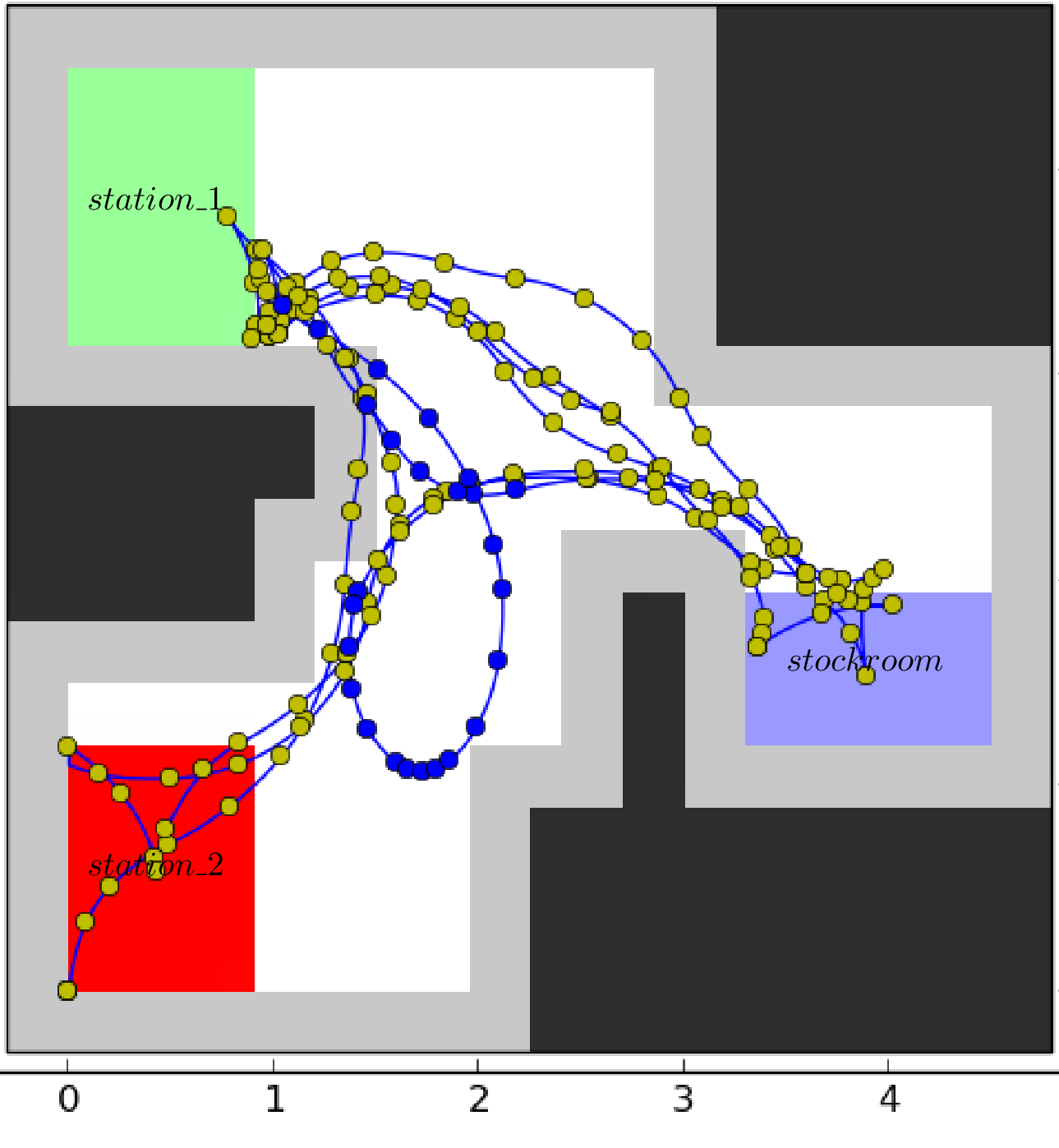}
}
\caption{Continuous trajectories for the nonlinear unicycle abstraction in a $5\times 5$ workspace, initialized at the lower-left corner of the workspace.  Regions $station\_1$, $station\_2$ and $stockroom$ are shown in green, red, and blue, respectively.  Dots along the trajectory indicate position at each discrete time step (0.35 seconds).  Color indicates the state of the environment (red: $s1\_occupied$; blue: $s2\_occupied$).  \protect\subref{f:traj1} shows a trajectory when the $s1\_occupied$ sensor is activated.  \protect\subref{f:traj2} shows a trajectory when the $s2\_occupied$ sensor is activated. }
\label{f:traj_main_examp}
\end{figure*}

The general specification is realizable, producing the controller pictured in Figure~\ref{f:aut_main_examp}; however the specification $\varphi^{abs}$ (with respect to the unicycle model) is unrealizable.  With the approach in Algorithm~\ref{a:refine}, we compose revisions that render $\varphi^{mod}$ realizable.  After a counterstrategy is synthesized, revisions are found for a total of 2040 states in the counterstrategy (taking 1020 seconds to synthesize on a laptop PC with a dual-core processor and 8GB memory).  A metric for these revisions is generated and the user is prompted with the following: 
\begin{flushleft}
\begin{small}
\texttt{Deadlock revisions found.  \\When within 1.32 m of $station\_1$, never set environment variable $s1\_occupied$ to \true. \\Accept? (y/n)}
\end{small}
\end{flushleft}
Note that, as our configuration space consists of variables of mixed units, the norm computed in~\eqref{e:furthestStates} has been projected onto the Cartesian plane.  A second prompt is given: 
\begin{flushleft}
\begin{small}
\texttt{When within 1.44 meters of $station\_2$, never set environment variable $s2\_occupied$ to \true. \\Accept? (y/n)}
\end{small}
\end{flushleft}
At this point, should the user accept both revisions, a new counterstrategy is synthesized containing no deadlock states.  The user is prompted again:
\begin{flushleft}
\begin{small}
\texttt{Livelock revisions found.  \\Accept? (y/n)}
\end{small}
\end{flushleft}
This time, the specification is realizable if the user accepts and the resulting execution for the controller is as shown in Figure~\ref{f:traj_main_examp}.  The trajectories pictured in the figure represent evolutions of the continuous nonlinear system when commanded by the synthesized controller.  Forward integration is applied to solve the equations of motion using an integration step size of $0.001$ sec.  Here $\varepsilon=0.3$, and the regions are inflated to the extent indicated by the gray border.  Note that the system in the figure infinitely often visits the three regions and is able to react to a change in the environment.  In Figure~\ref{f:traj1}, the system avoids the region $station\_1$ when $s1\_occupied$ becomes \true, since this does not happen within 1.32 m of $station\_1$.  A similar result is seen in Figure~\ref{f:traj2}.  These behaviors are consistent with the intended behaviors encoded by the specification in Example~\ref{ex:main_examp}.  
%Note that there is conservatism built into the result.  One of the reasons for this conservatism is that, at each configuration, the robot is subject to non-determinism.  The result is that each state for which there are no actions where all possible successors remain in the set of winning states is itself not winning.  In actuality, the continuous trajectory may not exhibit any such behaviors.

\section{Conclusions} \label{sec:conclusion}
In this paper, we have described an automatic approach for synthesizing controllers for dynamical systems based on general specifications that are agnostic to the dynamics.  We focus on the case where such specifications are transformed based on the discrete abstraction for a particular robot, and develop a framework for revising the specification in the case when the dynamics render the task unrealizable.  We introduce a method for automatically generating a set of LTL formulas that, when added to the original specification, render it realizable.  The approach features a mechanism for providing feedback to the user, giving him or her the freedom to accept or reject any such proposed formula.  To facilitate interpretation of such formulas, the feedback given to the user is metric information that encapulates the required restrictions on the environment and system behaviors due to the suggested revisions.  Future work includes providing a means for suggesting a richer set of possible revisions to give as feedback to the user, thereby offering him or her a multiplicity of possible options to apply (e.g. trading off modifying the robot's behavior vs. restricting the environment).  Such an extension will involve mining more complex formulas from the synthesis game and automatically translating such formulas into easy-to-understand explanations.

\section*{Acknowledgement}
The authors thank Salar Moarref, Ufuk Topcu, and Rajeev Alur for insightful discussions relating to synthesis of counterstrategy-based environment revisions.

%\bibliographystyle{harvard}
%\bibliography{References}

%\cite{pagedas:flexible}, ccccccc\citep{prat.flick.ea:biodiversity}, 
%\citet{filamin,21st,american} 
%\citet{bibliographic, cancer-pain, compendium,hypertension,mesh,outreach}
%\citep{whos,abend.kulish:psychoanalytic,ahrar.madoff.ea:development,breedlove.schorfheide:adolescent,christensen.oppacher:analysis}

%\bibliographystyle{vancouver}
%\bibliography{References}

\end{document}